\documentclass[twoside,11pt]{article}

\usepackage{blindtext}
\usepackage{amsmath}
\usepackage{booktabs}
\usepackage{multirow}

\usepackage[pdftex]{graphicx}
\usepackage{amsmath}
\usepackage{amssymb}
\usepackage{algorithm}
\usepackage{algorithmicx}
\usepackage{algpseudocode}
\usepackage{booktabs}
\usepackage{diagbox} 

\usepackage{caption}
\usepackage{subcaption}
\usepackage{url}
\usepackage{tabularx}
\usepackage{mathtools}
\usepackage{array, booktabs}
\usepackage{graphicx}
\usepackage{xcolor}         

\usepackage{jmlr2e}
\usepackage{color}

\usepackage{lastpage}

\firstpageno{1}

\begin{document}

\title{Trainable Weight Averaging: Accelerating Training and Improving Generalization}

\author{\name Tao Li \email li.tao@sjtu.edu.cn \\
\name Zhehao Huang \email kinght\_h@sjtu.edu.cn \\
\name Yingwen Wu \email yingwen\_wu@sjtu.edu.cn \\
\name Zhengbao He \email lstefanie@sjtu.edu.cn \\
       \addr 
       Department of Automation,
       Shanghai Jiao Tong University\\
\name Qinghua Tao \email qinghua.tao@esat.kuleuven.be \\
\addr
ESAT-STADIUS, 
KU Leuven \\ 
\name Xiaolin Huang \email xiaolinhuang@sjtu.edu.cn \\
       \addr 
       Department of Automation, 
       Shanghai Jiao Tong University\\
\name Chih-Jen Lin \email cjlin@csie.ntu.edu.tw \\
       \addr 
       Department of Computer Science,
       National Taiwan University \\
       Mohamed bin Zayed University of Artificial Intelligence \\
       }

\editor{My editor}

\maketitle

\begin{abstract}
    Weight averaging is a widely used technique for accelerating training and improving the generalization of deep neural networks (DNNs). 
    While existing approaches like stochastic weight averaging (SWA) rely on pre-set weighting schemes, they can be suboptimal when handling diverse weights.
    We introduce \emph{Trainable Weight Averaging} (TWA), a novel optimization method that operates within a reduced subspace spanned by candidate weights and learns optimal weighting coefficients through optimization.
    TWA offers greater flexibility and can be applied to different training scenarios.
    For large-scale applications, we develop a distributed training framework that combines parallel computation with low-bit compression for the projection matrix, effectively managing memory and computational demands.
    TWA can be implemented using either training data (TWA-t) or validation data (TWA-v), with the latter providing more effective averaging.
    Extensive experiments showcase TWA's advantages: 
    (i) it consistently outperforms SWA in generalization performance and flexibility,
    (ii) when applied during early training, it reduces training time by over 40\% on CIFAR datasets and 30\% on ImageNet while maintaining comparable performance,
    and (iii) during fine-tuning, it significantly enhances generalization by weighted averaging of model checkpoints.
    In summary, we present an efficient and effective framework for trainable weight averaging.
    The code is 
    available at \url{https://github.com/nblt/TWA}.
\end{abstract}


\section{Introduction}
Weight averaging is a widely used technique for accelerating training and improving the generalization performance of deep neural networks (DNNs)~\citep{izmailov2018averaging,gupta2019stochastic,yang2019swalp,kaddour2022stop,wortsman2022model}. 
It exploits the linear mode connectivity~\citep{frankle2020linear} in DNNs' loss landscapes, effectively mimicking model ensemble performance within a single weight space. Due to its simplicity and effectiveness, weight averaging has gained significant attention from practitioners.

In SWA~\citep{izmailov2018averaging}, weight averaging is given by $\boldsymbol{w}_{\mathrm{swa}}=\sum_{i=1}^{n}\frac{1}{n}\boldsymbol{w}_{i} $, where $n$ candidate solutions of the network collected at the tail stage of training are equally averaged. This strategy has been shown to be effective in enhancing generalization ability.
Other improved strategies include selective averaging methods, such as greedy soup~\citep{wortsman2022model} and latest weight averaging (LAWA)~\citep{kaddour2022stop}, as well as leveraging pre-set strategies like exponential moving average (EMA).
However, applying equal or pre-set weighting coefficients to candidate model weights can be inappropriate, especially when weights differ significantly, such as those sampled from different training configurations. This can potentially result in suboptimal performance.

In this paper, we study weight averaging with learnable coefficients.
Let us regard each model weight as a basis. When we train the coefficients without considering the dependence of different weights, the number of model weights used for averaging is roughly equal to the dimension of the optimization space. It has been reported that for deep models, the training dynamics happen in a low-dimensional subspace~\citep{gur2018gradient,li2022low} and crucial DNN mode-connectivity patterns emerge early in training~\citep{you2019drawing,frankle2020linear}. Therefore, optimizing the weighting coefficients provides great flexibility and can lead to improvements compared to equal averaging. This is especially important for applying weight averaging in the early stage when the parameters have not been well trained. 
These findings suggest a promising direction: effectively utilizing these early explorations may enable rapid composition of the final solution while maintaining high accuracy. 
Since many model weights are involved, optimizing the weighted coefficients also faces practical challenges in both memory and computation. Efficiency is an obstacle that must be solved when dealing with large-scale model training~\citep{wortsman2022model}.

In this paper, we delve into efficient and effective weighted averaging of DNNs. 
We introduce \emph{Trainable Weight Averaging} (TWA), a novel approach that enables explicit, trainable adjustments to the weighting coefficients.
To facilitate efficient optimization, we reframe the problem as a subspace training task: regarding each weight as a point in the full parameter space, we construct a subspace containing the weight points to be averaged.
By optimizing in this subspace, we can adaptively search for a good set of weighting coefficients.
This scheme, which involves gradient projection onto the subspace, is computationally efficient and can be accelerated by GPUs.
To cope with the challenge of serve 
memory burden when averaging with multiple weights~\citep{wortsman2022model}, we develop an efficient distributed subspace training scheme that facilitates parallel processing across multiple nodes. 
This approach efficiently handles large-scale problems by evenly distributing memory and computational loads across nodes and can be seamlessly integrated with existing distributed training frameworks, such as Distributed Data Parallel (DDP)~\citep{li2020pytorch}.
Furthermore, 
to enhance averaging efficiency,
we introduce layer-wise processing to better utilize the model's structure information and find that the matrix associated with gradient projection can be quantized to low-bit representations (e.g., 4 bits) without performance degradation, further reducing memory usage.  Consequently, TWA enables efficient subspace training with minimal additional memory overhead, making it scalable for averaging large models.

By exploiting the low-dimensional nature of the subspace, we reduce the degrees of freedom from millions or billions to just dozens or hundreds. This reduction allows us to train with a relatively small number of samples. We, therefore, propose optimizing these coefficients using a small held-out validation set, leading to our variant TWA-v, in contrast to the version using training data, TWA-t. Since validation data are unseen in the training set, they could serve as new data to evaluate generalization capability and have played a crucial role in machine learning. For example, we use the validation set to assess the model checkpoints during the training and select the best one. We now have new ways to better utilize these model checkpoints by appropriately averaging with TWA-v, resulting in improved overall performance.



We demonstrate TWA's effectiveness in two scenarios through extensive experiments across model architectures and tasks:
1)  accelerating training by averaging historical solutions during the head stage of training, and 2) improving generalization by averaging solutions fine-tuned from single or multiple fine-tuning configurations.
Our results show over 40\% and 30\% reduction in training epochs on CIFAR-100 and ImageNet respectively, while maintaining or improving performance (Tables~\ref{table-cifar}, \ref{table-imagenet}).
We also demonstrate that TWA can better utilize the fine-tuned models, reducing the number of models required by 4x for reaching comparable performance compared to previous methods (Table~\ref{tab:finetune-ablation}).
Interestingly, we also find that TWA-v exhibits particular effectiveness with transformer architectures.


In summary, our contributions can be summarized as follows:
\begin{enumerate}
    \item We propose \emph{Trainable Weight Averaging} (TWA), an efficient approach that enables the averaging of weights with layer-wise learnable coefficients to accelerate training and improve the generalization of DNNs.
    
    \item We design an efficient scheme to handle large-scale problems via subspace training, enabling multi-node parallel training by evenly distributing the memory and computation burden into different nodes. We also devise a compression strategy to further reduce the memory footprint, facilitating effective coefficient optimization.
    
    \item Based on the fact that the number of learnable variables is small, we propose TWA-v that optimizes the coefficients with a small held-out validation set to enable more efficient and effective averaging. Additionally, we discover that TWA-v is particularly effective for transformer-based architectures.
    
    \item We demonstrate the efficiency and effectiveness of TWA through extensive experiments involving various architectures (e.g., CNNs, ViTs, and GPT-2), different tasks (e.g., image classification, machine translation, and language modeling), and multiple training scenarios (from scratch training to fine-tuning) using different optimizers (SGD and Adam).
    
\end{enumerate}


\paragraph{Comparison with our conference work.}
The content of this paper builds upon the previous ICLR 2023 conference version~\citep{li2023trainable} and includes the following substantial enhancements:
1) \emph{Refined Subspace Construction}: 
We have refined the subspace construction by incorporating simple decentralization and normalization techniques. This allows us to avoid heavy and unstable numerical operations in the vector orthogonalization process and makes it more amenable to parallel processing. Additionally, we introduce layer-wise processing to better leverage the model's structural information~(Section~\ref{sec:subspace-construction}).
2) \emph{Quantization of the Projection Matrix}: We have developed a low-bit compression method that reduces the memory overhead associated with gradient projection by 8x, facilitating efficient large-scale training~(Section~\ref{sec:training-speed});
3) \emph{Validation Supervised Optimization}: We propose using a small held-out validation set to supervise the optimization of the weighting coefficients, leading to TWA-v. This introduces a new approach to utilizing validation sets during training by learning weighted averages of historical checkpoints, 
rather than selecting a single best model by checking the validation performance after each training epoch (Section~\ref{sec:twa}).
4) \emph{Expanded Evaluation Scope}: 
We have extended the TWA's application to fine-tuning tasks (Section~\ref{sec:finetune}), improving fine-tuning performance by averaging weights fine-tuned from different training configurations. We have also conducted experiments across a broader range of tasks and model architectures to demonstrate the efficiency and effectiveness of our approach. Specifically, we evaluate TWA on machine translation (Section~\ref{sec:machine_translation}) and language modeling (Section~\ref{sec:language_modeling}) tasks and on ViT (Section~\ref{sec:img-classification}) and GPTs (Section~\ref{sec:language_modeling}).
5) \emph{New Findings}: Our investigations reveal that TWA-v is particularly effective when applied to transformer-based architectures, which are characterized by minimal inductive bias. This highlights the great potential of TWA-v since transformers are the core architecture for modern AI applications, such as large language models.

\section{Related Work}
\label{sec:related works}
Training neural networks in subspaces has recently become an interesting topic, garnering considerable interest from researchers \citep{vinyals2012krylov, gur2018gradient, tuddenham2020quasi}.
The pioneering work \citep{li2018measuring} first proposed training neural networks in a reduced random subspace to measure the intrinsic dimension of loss objectives. The following work \citep{gressmann2020improving} improved the training performance of random bases by considering the layer-wise structure and re-drawing the random bases at each step. 
Instead of utilizing random bases, \cite{li2022low} proposed a low-dimensional trajectory hypothesis and extracted the subspaces from historical training dynamics, dramatically improving the dimensionality efficiency. 
Then \cite{li2022subspace} applied subspace training to adversarial training, effectively addressing the existing catastrophic and robust overfittings, and thereby significantly improving the model robustness performance. 
In this paper, we reframe trainable weight averaging as a subspace training problem and develop an efficient training scheme that enables multi-node parallel training for large-scale problems, incorporating improved subspace construction procedures.


A lot of efforts have been made to speed up DNNs' training \citep{shen2023efficient}. 
Apart from the well-known optimization methods on adaptive learning rates, e.g., Adam \citep{kingma2014adam} or accelerated schemes, e.g., Nesterov momentum \citep{nesterov1983method},
\cite{zhang2019lookahead} proposed LookAhead optimizer that utilizes the search direction generated by another ``fast'' optimizer, achieving faster convergence and better learning stability.
\cite{goyal2017accurate} adapted a large mini-batch to speed up the training and introduced a scaling rule for adjusting the learning rates. 
\cite{you2017large, You2020Large} proposed a layer-wise adaptive learning rate to further scale the batch size and shorten the training time.
\cite{gupta2019stochastic} proposed to use large mini-batches to compute an approximate solution quickly and then refined it by averaging the weights of multiple
models computed independently and in parallel to accelerate the training.
In this paper, we improve the DNNs' training efficiency by sufficiently utilizing the historical solutions during the training and conducting training in a subspace with substantially reduced dimensions. In this way, we significantly reduce the required training epochs.


\begin{table}[!t]
    \centering
    \caption{Notations.}
    \vspace{2mm}\small
    \label{tab:notation}
    \begin{tabular}{cc}
    \toprule
         $D$ &the number of model parameters \\
         $L$ &the number of model layers \\
         $l$ &the $l$-th layer \\
         $n$ &the number of model weights\\
         $\boldsymbol{w}$ &model weights \\
         $\boldsymbol{g}$ &model gradient \\
         $\alpha$ &weighting coefficient \\
         $\boldsymbol{e}$ &base vector \\
         $\boldsymbol{\beta}$ &the coefficients for the base vectors\\
         $k$ &the number of nodes for distributed training \\
         $\mathcal{L}$ &loss objective \\
         $\mathcal{S}$ & datasets \\
         $\mathcal{B}$ &data batch \\
         $\boldsymbol{x}$ & data input \\         
         $\boldsymbol{y}$ & data label \\
         ${\boldsymbol{P}}$ &projection matrix\\
         $\Tilde{\boldsymbol{P}}$ &quantized projection matrix\\
         $B$ &quantization bits \\
         $a$ &quantization scaling factor \\
         $b$ &quantization zero factor \\
         $U$ &the subspace containing the weights \\
         $\tilde{U}$ &the subspace containing the weights (layer-wise) \\
         $\eta$ &learning rate \\
    \bottomrule
    \end{tabular}
\end{table}


Next, we review previous weight-averaging strategies.

\paragraph{Preliminary.}
In this paper, we consider a neural network function $f(\boldsymbol{w},\boldsymbol{x})$ parameterized by weights $\boldsymbol{w}$ with input $\boldsymbol{x}$. The training loss defined over a pair of data point $(\boldsymbol{x}_i,\boldsymbol{y}_i)$ is denoted as $\displaystyle \mathcal{L}(f(\boldsymbol{w};\boldsymbol{x}_i),\boldsymbol{y}_i)$ (shortened to $\mathcal{L}_i(\boldsymbol{w})$). Given a dataset $ \displaystyle \mathcal{S}^c=\left \{ (\boldsymbol{x}_i,\boldsymbol{y}_i) \right \}_{i=1}^{|\mathcal{S}^c|}$ drawn from the data distribution $\mathcal{D}$ with i.i.d. condition, the empirical loss is defined as $\mathcal{L}^c(\boldsymbol{w})=\frac{1}{|\mathcal{S}^c|}\sum_{i=1}^{|\mathcal{S}^c|} \mathcal{L}_i(\boldsymbol{w})$, where $c$ can be ``train'', ``val'', or ``test''.
Note that in this paper, we represent the model's weights as a vector, i.e., $\boldsymbol{w} \in \mathbb{R}^{D}$, where $D$ is the number of model parameters. The main notations used in this paper are listed in Table~\ref{tab:notation}.

In SWA~\citep{gupta2019stochastic}, weight averaging is simply given by
\begin{align}
\boldsymbol{w}_{\mathrm{swa}}=\sum_{i=1}^{n} \frac{1}{n} 
 \boldsymbol{w}_{i},
\end{align}
where $n$ solutions of the network are equally weighted. 
This strategy has proven effective when all the solutions are already well-optimized, such as in the tail stage of training. 
However, as a static averaging approach, it is susceptible to suboptimal solutions and may not be suitable for more general scenarios, such as during the head stage of training when model weights are rapidly evolving or for diverse models fine-tuned with different training configurations.

Recently, LAWA~\citep{kaddour2022stop} proposed to apply SWA to a consecutive segment of the most recent weight checkpoints along the training trajectory, i.e.,
\begin{align}
\boldsymbol{w}_{\mathrm{lawa}}=\sum_{i=n-t+1}^{n} \frac{1}{t} 
 \boldsymbol{w}_{i},
\end{align}
where $t$ is the averaging horizon that has to be pre-set. LAWA has shown effectiveness in accelerating the training of DNNs~\citep{kaddour2022stop, sanyal2023early}.

Besides, EMA~\citep{polyak1992acceleration} averages the model weights using an exponentially decayed factor $\gamma$, i.e.,
\begin{align}
\boldsymbol{w}_{\mathrm{ema}}=\sum_{i=1}^n \frac{1-\gamma}{1-\gamma^n} \gamma^{n-i}\boldsymbol{w}_i.
\end{align}
EMA typically averages model weights at each iteration.


Greedy soup~\citep{wortsman2022model} improves upon SWA by selectively averaging a subset of models. Specifically, it first sorts the fine-tuned models based on their validation accuracies and then sequentially adds models to the soup if doing so improves the validation performance. Greedy soup has been shown to empirically outperform SWA and is widely adopted by practitioners~\citep{rame2022diverse,croce2023seasoning}.

\begin{table*}[!t]
    \centering
    \caption{A comparison of related weight averaging methods from the perspective of the solution set. ``\#Dimension'' denotes the dimensionality of the solution space. ``Inference cost''
refers to the memory and computational requirements during inference relative to those of a single model. ``\#Dimension=0'' indicates that the method employs a fixed averaging strategy.}
    \footnotesize
    \label{tab:comparison}
    \renewcommand{\arraystretch}{1.2} 
    \resizebox{\linewidth}{!}{
    \begin{tabular}{lccccc}
    \toprule
         Method &Solution set &\#Dimension  &Inference cost  \\
    \midrule
    SWA 
    &$\left \{\sum_{i=1}^n \frac{1}{n} \boldsymbol{w}_i\right \}$ &0  &$\mathcal{O}(1)$ 
    \\
    LAWA 
    &$\left \{\sum_{i=n-t+1}^n \frac{1}{t}\boldsymbol{w}_i\right \}$ &0 &$\mathcal{O}(1)$
    \\ 
    EMA &$\left \{\sum_{i=1}^n \frac{1-\gamma}{1-\gamma^n} \gamma^{n-i}\boldsymbol{w}_i\right \}$ &0  &$\mathcal{O}(1)$\\
    Greedy soup 
     &$\left\{  \sum_{i=1}^t \frac{1}{t} \boldsymbol{w}_i~|~ \mathrm{for} ~~t \in [n], \text{with}~\texttt{ValAcc}(\boldsymbol{w}_i)~\text{decreasing} \right\} $ 
    &0    &{$\mathcal{O}(1)$}
    \\
    TWA (w/o layer-wise) &$\mathrm{span} \left \{ \boldsymbol{w}_i \right \}_{i=1}^n$ &$n$  &$\mathcal{O}(1)$  
     \\ 
    TWA 
    & ${\rm span}  \{ \boldsymbol{w}^{1}_i  \}_{i=1}^n   \oplus {\rm span}   \{ \boldsymbol{w}^{2}_i  \}_{i=1}^n  \cdots \oplus{{\rm span}}  \{ \boldsymbol{w}^{L}_i \}_{i=1}^n$
    &$nL$  &$\mathcal{O}(1)$  \\
    \midrule 
    Ensemble &$\left \{ [\boldsymbol{w}_1,\boldsymbol{w}_2,\cdots,\boldsymbol{w}_n] \right \}$  &0 &$\mathcal{O}(n)$\\
    \bottomrule
    \end{tabular}}
\end{table*}

\section{Trainable Weight Averaging}
\label{sec:twa}

In this paper, we propose \emph{trainable weight averaging} (TWA), a method that optimizes the weighting coefficients of different model weights to improve performance.
Given a set of weights $\left\{\boldsymbol{w}_i \right\}_{i=1}^n$, 
which can be sampled either from a single training trajectory or collected from multiple fine-tuning configurations, 
we aim to linearly combine them for creating an improved averaged solution. Specifically, the set of possible TWA solutions we considered, i.e.,  $\boldsymbol{w}_{\mathrm{twa}}$, can be represented as follows:
\begin{align}
    U = \left \{ \boldsymbol{w}_{\mathrm{twa}} ~~|~~  \boldsymbol{w}_{\mathrm{twa}}=\alpha_1 \boldsymbol{w}_{1} + \alpha_2  \boldsymbol{w}_{2} + \cdots + \alpha_n \boldsymbol{w}_{n},~  \alpha_i \in \mathbb{R}  \right \}.
    \label{equ:solution}
\end{align}
One can see that the solution set $U$ forms a linear space with dimension $n$.

Previous pre-set averaging strategies may fall short for complex weights, potentially leading to averaged solutions that are suboptimal within the subspace $U$.
In fact, the solutions obtained by these approaches can be viewed as specific points within $U$. 
TWA addresses this limitation by optimizing the weighting coefficients to achieve better performance. A comparison between TWA and previous averaging approaches is presented in Table \ref{tab:comparison}.

We now consider how to search for the optimal solution $\boldsymbol{w}^*_{\mathrm{twa}}$ within the subspace ${U}$. 
First, we will identify a set of bases $\{\boldsymbol{e}_i\}_{i=1}^n$ to support the solution space such that ${U}=\left\{ \beta_1 \boldsymbol{e}_{1} + \beta_2  \boldsymbol{e}_{2} + \cdots + \beta_n \boldsymbol{e}_{n} ~~|~~ \beta_i \in \mathbb{R} \right \}$, where $\{\beta_i\}_{i=1}^n$ are the coefficients to be optimized.
Rather than using $\{{\boldsymbol{w}_i}\}_{i=1}^n$ directly as bases, we would decouple them to $\{\boldsymbol{e}_i\}_{i=1}^n$
since the weight vectors $\{{\boldsymbol{w}_i}\}_{i=1}^n$ are often highly correlated, 
particularly when fine-tuned from the same pre-trained model. Such correlations complicate optimization and can degrade the training performance, as shown in Section~\ref{sec:decouple}.
Then we search for $\boldsymbol{w}^*_{\mathrm{twa}}$ in ${A}$ by optimizing $\beta_i$ with the following problem,
\begin{align}
\begin{split}
    \min_{\beta_1, \beta_2, \cdots, \beta_n} \quad 
    &\sum_{(\boldsymbol{x}_j, \boldsymbol{y}_j)\in \boldsymbol{S}} 
    \mathcal{L}  \left ( f \left (\boldsymbol{w}_{\rm twa}; \boldsymbol{x}_j \right ), \boldsymbol{y}_j \right ) 
    +
    \frac{\lambda}{2}  \sum_{i=1}^{n}  \beta_i^2  \\
    \rm{s.t.}\qquad &\boldsymbol{w}_{\mathrm{twa}}= \beta_1 \boldsymbol{e}_{1} + \beta_2  \boldsymbol{e}_{2} + \cdots + \beta_n \boldsymbol{e}_{n}.
    \label{equ-target}
    \end{split}
\end{align} 
The second term serves as a regularization for $\beta_i$ with a coefficient $\lambda>0$. 

There are two choices for using the available data to optimize Eqn.~(\ref{equ-target}): 
\begin{itemize}
    \item Using the training data, i.e., $\boldsymbol{S} = \boldsymbol{S}^{\text{train}}$, while still leveraging the validation data $\boldsymbol{S}^{\text{val}}$ to select the best-performing model during training after each training epoch.
    \item Using the validation data directly, i.e., \(\boldsymbol{S} = \boldsymbol{S}^{\text{val}}\),
to optimize the weighting coefficients and select the final model after the training is complete.
\end{itemize}
The latter approach can be more efficient and effective:
the rationale behind this choice is that since we only need to optimize a small number of coefficients, a small held-out validation set $\boldsymbol{S}^{\text{val}}$ is sufficient. Given that the weights $\{\boldsymbol{w}_i\}_{i=1}^n$ are unseen to $\boldsymbol{S}^{\text{val}}$, if $\boldsymbol{w}_{\rm twa}$ performs well on the validation set, it is expected to generalize effectively during testing.
Additionally, the validation set can be typically much smaller than the training set, which helps reduce training costs.
We denote this validation-based approach as TWA-v, distinguishing it from the training-based approach denoted as TWA-t.
This introduces a novel use of validation data by learning optimal weighting coefficients to combine multiple models, rather than selecting a single best-performing model.

\section{An Efficient Training Scheme}
\label{sec:scheme}

A straightforward approach to optimizing the coefficients $\{\beta_i\}_{i=1}^n$ is to construct a computational graph that links the bases $\left\{\boldsymbol{e}_i\right\}_{i=1}^n$ and coefficients to the model weights $\boldsymbol{w}$, enabling gradient descent on $\{\beta_i\}_{i=1}^n$ through standard backpropagation in an end-to-end manner. However, as noted by \cite{wortsman2022model}, this approach results in a large computational graph, introducing excessive additional memory overhead and becoming prohibitive as the model size and number of weights increase.

In this paper, we propose a simpler and more efficient method for optimizing the coefficients $\beta_i$ without constructing an additional computational graph, by leveraging subspace training. 
We notice that there exists a mapping between the coefficient space $\{ \beta_i \}_{i=1}^n \in \mathbb{R}^n$ and the parameter space $\mathbb{R}^D$:
\begin{align}
    \boldsymbol{w}=\sum_{i=1}^n \beta_i \boldsymbol{e}_i,
\end{align}
This implies that each set of coefficients $\{\beta_i\}_{i=1}^n$ uniquely corresponds to a point $\boldsymbol{w}$ in the parameter space $\mathbb{R}^D$, thereby forming a subspace of dimensionality $n$. As a result, we can optimize the coefficients $\{\beta_i\}_{i=1}^n$ by optimizing $\boldsymbol{w}$ within this subspace, which is more efficient than performing direct optimization through a computational graph.

We then present our subspace training scheme for efficient trainable weight averaging, which consists of subspace construction and training. To further enhance performance and efficiency, we introduce a layer-wise extension to incorporate model structure, develop a multi-node distributed training approach for large-scale problems, and quantize the projection matrix to reduce memory overhead. Before detailing the training procedure, we first discuss the challenges of averaging multiple model weights.
\begin{itemize}
    \item 
\textbf{Computation.}
As the model size and the number of models grow, the costs of both subspace construction and subspace training increase accordingly.
\item \textbf{Memory.}
Subspace training requires storing a matrix for projecting gradients onto the subspace, which can be challenging in large-scale scenarios where the matrix may exceed a single GPU's memory capacity.
    \item \textbf{Balance load.}
    Both subspace construction and training are best performed on GPUs, with the computational load distributed across multiple nodes when available. This significantly reduces training time and improves the scalability of the algorithm.
\end{itemize}

\subsection{Subspace Construction}
\label{sec:subspace-construction}
Given $n$ candidate weights $\left \{ \boldsymbol{w}_i \right \}_{i=1}^n$, where $\boldsymbol{w}_i \in \mathbb{R}^D$, we aim to construct a subspace covering these weight points. 
To achieve this, we first identify a set of bases to represent the subspace. 
Since the weights are often interdependent---particularly when sampled from a single training trajectory, we avoid directly using them as bases. Leveraging the fact that the high dimensionality of the weight space, which naturally, encourages independence,
and the weights originate from a single linear mode\footnote{
The concept of linear mode connectivity, as defined by \citet{frankle2020linear}, refers to the property where, for two sets of weights $\boldsymbol{w}_1$ and $\boldsymbol{w}_2$, their linear interpolation $\boldsymbol{w}(\alpha) = (1 - \alpha) \boldsymbol{w}_1 + \alpha \boldsymbol{w}_2$ maintains good performance for all $\alpha \in [0, 1]$.
}, we propose to decouple them by performing simple decentralization and normalization, as follows:
\begin{align}
\begin{split}
    { \boldsymbol{\overline w}} &= \frac{1}{n} \sum_{i=1}^n \boldsymbol{w}_i, \\
    \boldsymbol{e}_i &= \frac{\boldsymbol{w}_i - {\boldsymbol{\overline w}}}{\| \boldsymbol{w}_i - {\boldsymbol{\overline w}} \|_2}.
\end{split}
\label{equ:extraction}
\end{align}
This results in a set of basis vectors $\left \{ \boldsymbol{e}_i \right \}_{i=1}^n$.
Next, we optimize the neural network within the subspace $U=\left \{ \sum_i \beta_i \boldsymbol{e}_i \right \}$. 
We represent a weight point in this subspace using the variable $\boldsymbol{\beta}=\left [\beta_1,\beta_2,\cdots,\beta_n \right ]^\top$. It is important to note that the dimension of $U$ is $n$, the number of models, which is much smaller than $D$, the dimension of the original parameter space.



Another approach to decouple the model weights $\left \{ \boldsymbol{w}_i \right \}_{i=1}^n$ is to orthogonalize them using Schmidt orthogonalization~\citep{li2023trainable}.
However, this sequential process can be computationally intensive,
potentially compromising the accuracy of subspace basis estimation.
In contrast, our decoupling method, which involves only decentralization and normalization of the weight vectors without intensive numerical operations, does not enforce strict orthogonalization. Our experiments show that this relaxation does not significantly impact the subspace training performance. In fact, this simplified strategy effectively decouples the weight vectors while being readily applicable to distributed training, resulting in more efficient subspace construction.
We will compare this in Section \ref{sec:subspace-extraction}.

\begin{figure*}[!t]
\begin{center}
\includegraphics[width=0.85\linewidth]{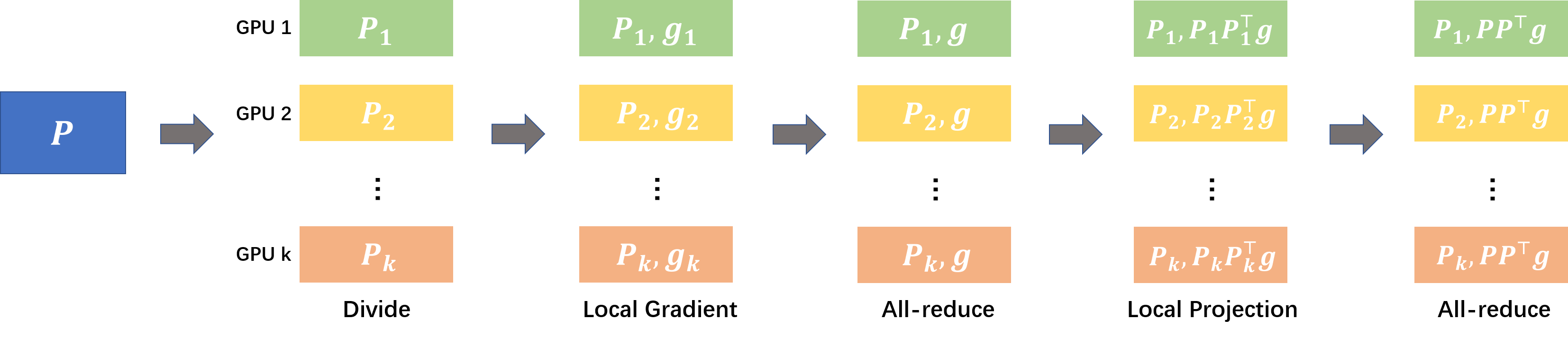}
\caption{\textbf{An efficient parallel scheme for subspace training.} Suppose there $k$ GPUs available  for distributed training. We begin by uniformly partitioning $\boldsymbol{P}$ into $k$ sub-matrices, i.e., $[\boldsymbol{P}_1, \boldsymbol{P}_2, \cdots, \boldsymbol{P}_k]$, with each GPU storing one sub-matrix. First, we synchronize the local gradient $\boldsymbol{g}_i$ to obtain global gradient $\boldsymbol{g}_i$ using an all-reduce operation. Next, each node computes the local projected gradient $\boldsymbol{P_i}\boldsymbol{P_i}^\top \boldsymbol{g}$ independently. Finally, we perform a second all-reduce operation to obtain the global projected gradient $\boldsymbol{P}\boldsymbol{P}^\top\boldsymbol{g}$.
In this way, we evenly distribute the memory and computation burden across all nodes.
}
\label{fig:scheme}
\end{center}
\end{figure*}

\subsection{Subspace Training}
\label{sec:subspace_training}
We then consider how to optimize the variable $\boldsymbol{\beta}$ for subspace training. Let $\boldsymbol{P}=\left [ \boldsymbol{e}_1, \boldsymbol{e}_2,\cdots, \boldsymbol{e}_n \right ]$. We parameterize the model weights in subspace as $\boldsymbol{w}_{\rm twa}=\boldsymbol{P}\boldsymbol{\beta} \in U$. Then the optimization target in Eqn.~(\ref{equ-target}) can be reformed as follows:
\begin{align}
\begin{split}
     \min_{\boldsymbol{\beta}} \quad 
    & \sum_{j=1}^m
    \mathcal{L}  \left ( f \left (\boldsymbol{w}_{\rm twa}; \boldsymbol{x}_j \right ), \boldsymbol{y}_j \right ) + \frac{\lambda}{2} \boldsymbol{\beta}^\top\boldsymbol{\beta}
    \\
    \rm{s.t.}\quad &\boldsymbol{w}_{\mathrm{twa}}= \boldsymbol{P}\boldsymbol{\beta}.
    \end{split}
\end{align}
The gradient of $ \mathcal{L}(\boldsymbol{w}_{\rm twa})$ w.r.t. to the variables $\boldsymbol{\beta}$ can be derived using the chain rule:
\begin{align}
    \frac{\partial \mathcal{L}(\boldsymbol{w}_{\rm twa})}{\partial \boldsymbol{\beta}}=
    \frac{\partial \boldsymbol{w}_{\rm twa}}{\partial \boldsymbol{\beta}} \frac{\partial \mathcal{L}(\boldsymbol{w}_{\rm twa})}{\partial \boldsymbol{w}_{\rm twa}} =\boldsymbol{P}^\top \frac{\partial \mathcal{L}(\boldsymbol{w}_{\rm twa})}{\partial \boldsymbol{w}_{\rm twa}}.
\end{align}
Thus, we can first calculate the model gradient $\displaystyle \boldsymbol{g}={\partial \mathcal{L}(\boldsymbol{w}_{\rm twa})}/{\partial \boldsymbol{w}_{\rm twa}}$ through standard forward and backward propagation, and then project the gradient onto the bases $\boldsymbol{P}$ to obtain the gradient w.r.t. $\boldsymbol{\beta}$. 
This projection ensures that the actual update to the model weights remains within the subspace $U$.
Together with the regularization, 
the gradient descent update rule for $\boldsymbol{\beta}$  is then given by:
\begin{align}
     \boldsymbol{\beta}^{(t+1)}=\boldsymbol{\beta}^{(t)}-\eta \left( \boldsymbol{P}^\top \frac{\partial \mathcal{L}(\boldsymbol{w}_{\rm twa}^{(t)})}{\partial \boldsymbol{w}_{\rm twa}} + \lambda \boldsymbol{\beta}^{(t)} \right),
     \label{equ:update}
\end{align}
where $\eta$ is the learning rate. 
For the corresponding TWA solution, we have:
\begin{align}
\boldsymbol{w}_{\rm twa}^{(t+1)}=(1-\eta \lambda)\boldsymbol{w}_{\rm twa}^{(t)}-\eta \boldsymbol{P} \boldsymbol{P}^\top \frac{\partial \mathcal{L}(\boldsymbol{w}_{\rm twa}^{(t)})}{\partial \boldsymbol{w}_{\rm twa}}.
\end{align}
Note that the gradient projection is achieved by applying the projection matrix $\boldsymbol{P}\boldsymbol{P}^\top$.


Here subspace training involves an additional gradient projection operation in the optimization step, compared to regular training.
This operation is a matrix multiplication, which can be efficiently executed on GPUs. As a result, the training speed of subspace training can be nearly the same as that of regular training. We compare the training speed of the two methods in detail in Section \ref{sec:training-speed}, where we find that TWA training is only slightly slower than regular training.



\subsection{Initialization}
An important consideration in TWA training is the choice of initialization. In this paper, we use the SWA solution as the initialization, i.e., $\boldsymbol{w}_{\rm twa}^{(0)} = \sum_{i=1}^n \frac{1}{n} \boldsymbol{w}_i$. This approach is advantageous because SWA typically delivers good performance, reducing the effort required for TWA training, and it avoids additional computation to evaluate the performance of individual model $\boldsymbol{w}_i$.

In practice, we set $\boldsymbol{\beta}^{(0)}=\mathbf{0}$ and use the following relation relationship:
\begin{align}
    \boldsymbol{w}^{(t)}_{\text{twa}} = {\boldsymbol{\overline w}} + {{{\boldsymbol{P}}}} \boldsymbol{\beta}^{(t)},
\end{align}
where ${\boldsymbol{\overline w}}=\sum_{i=1}^n \frac{1}{n} \boldsymbol{w}_i$ and $t$ is the number of iteration. During the training, we only need to maintain $\boldsymbol{w}^{(t)}_{\text{twa}}, {{{\boldsymbol{P}}}}, \boldsymbol{\beta}^{(t)}$ in the memory.

\subsection{Distributed Training Scheme}
During subspace optimization, TWA requires the storage of the projection matrix $\boldsymbol{P}$, which has a size of $\mathcal{O}(nD)$. This poses a storage challenge for large models, as $\boldsymbol{P}$ is ideally stored on the GPU to facilitate efficient matrix operations. As the model size and the number of weight points grow, storing $\boldsymbol{P}$ in a single GPU becomes impractical.

To cope with this, we design an efficient scheme with parallel distributed training to enable the following:
\begin{itemize}
    \item \textbf{Partitioning the memory burden of $\boldsymbol{P}$ into multiple GPUs.} We partition the projection matrix $\boldsymbol{P}$ into multiple submatrices, each of which can be stored in a separate GPU. This reduces the memory burden on each GPU and allows us to train larger models with scalability.
    \item \textbf{Efficient parallel computation of gradient projection.} We propose a distributed algorithm for computing the gradient projection in subspace optimization, which evenly distributes the computation load across multiple GPUs. It significantly reduces the time required for the gradient projection.
\end{itemize}


More specifically,
suppose that there are $k$ GPUs for multi-node parallel training. We will uniformly divide $\boldsymbol{P}$ into $k$ sub-matrices as $\boldsymbol{P}=[\boldsymbol{P}_1,\boldsymbol{P}_2,\cdots,\boldsymbol{P}_k]$. Each GPU stores a local sub-matrix $\boldsymbol{P}_i$ for $i=1, \cdots, k$. Recall that for an iteration in distributed training, each GPU computes a local gradient $\boldsymbol{g}_i$ and synchronizes it with other GPUs to obtain the global gradient through an efficient all-reduce operation \citep{rabenseifner2004optimization}. We mimic such a process for gradient projection.
First, we perform an all-reduce operation such that the local gradient $\boldsymbol{g}_i$ at each GPU is synchronized to $\frac{1}{k}\sum_{j=1}^k\boldsymbol{g}_j$. We then obtain local gradient projection $\frac{1}{k}\boldsymbol{P}_i\boldsymbol{P}_i^{\top}\big(\sum_{j=1}^k\boldsymbol{g}_j \big)$. Finally, we synchronize this to $\frac{1}{k}\sum_{i=1}^k\left[\boldsymbol{P}_i\boldsymbol{P}_i^{\top}(\sum_{j=1}^k\boldsymbol{g}_j)\right]$ with another all-reduce operation. The distributed approach is mathematically equivalent to the original gradient projection for the multiplication of the block matrix: 
\begin{align}
\begin{split}
\frac{1}{k}\boldsymbol{P}\boldsymbol{P}^{\top}{\bigg(}\sum_{j=1}^k\boldsymbol{g}_j{\bigg)}
=\frac{1}{k}
\begin{bmatrix}
\boldsymbol{P}_1,\boldsymbol{P}_2,\cdots,\boldsymbol{P}_k
\end{bmatrix}
\begin{bmatrix}
\boldsymbol{P}^{\top}_1 \\
\boldsymbol{P}^{\top}_2 \\
\vdots \\
\boldsymbol{P}^{\top}_k \\
\end{bmatrix}
{\bigg(}\sum_{j=1}^k\boldsymbol{g}_j{\bigg)}
=\frac{1}{k}\sum_{i=1}^k{\Bigg[}\boldsymbol{P}_i\boldsymbol{P}_i^{\top}{\bigg(}\sum_{j=1}^k\boldsymbol{g}_j{\bigg)}{\Bigg]}.
\end{split}
\end{align}
Note that in this way, the computation for matrix multiplication is also uniformly divided into different nodes.
We illustrate such a process in Figure \ref{fig:scheme}.


\subsection{Layer-wise Processing}
In the above, we have treated the model weights as a whole, where the layer-wise structure is not explored. It would be helpful to take into account the layer-wise information for more delicate subspace training. 

Suppose the network is composed of $L$ layers.
To incorporate layer-wise information, we propose constructing subspaces for each layer. We start by partitioning the concatenated weights into $L$ groups that correspond to each layer, i.e., $\boldsymbol{w}= [ \boldsymbol{w}^{1}, \boldsymbol{w}^{2}, \cdots, \boldsymbol{w}^{L} ]$. The augmented subspace we consider is 
\begin{align}
    \tilde{U}={\rm span}  \{ \boldsymbol{w}^{1}_i  \}_{i=1}^n   \oplus {\rm span}   \{ \boldsymbol{w}^{2}_i  \}_{i=1}^n  \cdots \oplus{{\rm span}}  \{ \boldsymbol{w}^{L}_i \}_{i=1}^n.
\end{align}
 We then normalize and decentralize each group of weights to produce a set of basis vectors corresponding to each layer following 
Eqn.~(\ref{equ:extraction}):
\begin{align}
        \boldsymbol{e}_i^l &= \frac{\boldsymbol{w}_i^l - {\boldsymbol{\overline w}}^l}{\| \boldsymbol{w}_i^l - {\boldsymbol{\overline w}}^l \|_2}.
        \label{equ:normalization}
\end{align}
Finally, we optimize the variables associated with each layer individually using the update rule as in Eqn.~(\ref{equ:update}). This layer-wise approach enables us to delicately fine-tune the subspaces for each layer, which can result in improved performance. 

In our main experiments, we process model weights layer-wise by default, as this approach yields better performance and enables effective quantization of the projection matrix, thereby reducing memory usage during gradient projection. We present an ablation study on layer-wise processing in Section~\ref{sec:layer}.

\subsection{Low-Bit Quantization of the Projection Matrix}
Since the weights to be averaged $\{\boldsymbol{w}_i\}_{i=1}^n$ are sampled from a single training trajectory or linear mode, it can be expected that the resulted projection matrix contains redundant information and can be further compressed.
To further reduce the memory burden introduced by the projection matrix $\boldsymbol{P}$, 
we resort to quantization, a widely used technique that is shown effective in reducing the memory burden and speeding up the training~\citep{shen2020q,dettmers2022gpt3,dettmers2024qlora,zhang2024q}.

\begin{table}[t]
    \centering
    \renewcommand{\arraystretch}{1.5} 
    \begin{tabular}{c|cc}
          \diagbox[width=5em,height=2.5em]{\small Nodes}{\small Bits} &32 &$B$  \\
         \hline
          $1$ &$4nD$ &$\frac{nBD}{8}+8L$ \\
         $k$ &$ 4 \displaystyle \lceil {n}/{k} \rceil D $ &$ \frac{\lceil {n}/{k} \rceil B D}{8} +8L$
    \end{tabular}
    \caption{A comparison of the memory required for the projection matrix $\boldsymbol{P}$ in TWA (in Bytes). Here, $n$ is the number  of models, $D$ is the number of model parameters, $k$ is the number of nodes to process, and $L$ is the number of layers. The memory cost of $\boldsymbol{P}$ can be reduced by both the number of nodes and the quantization bits.}
\label{tab:memory_projection}
\end{table}

We discover that the projection matrix $\boldsymbol{P}$ can be efficiently quantized into low-bit representation $\Tilde{\boldsymbol{P}}$. 
In this paper, we employ simple uniform min-max quantization, and more delicate methods can further improve the performance.
Mathematically, given the bit width $B$ and model layer number $l$, $\boldsymbol{P}$ is quantized into $\Tilde{\boldsymbol{P}}$ by computing:
\begin{align}
    \Tilde{\boldsymbol{P}}^{l} 
    =a^{l} \cdot \left \lfloor \frac{\boldsymbol{P}^{l}-b^{l}}{a^{l}} \right \rceil + b^{l},
    \label{equ:quantization}
\end{align}
where $a^l=(\max(\boldsymbol{P}^{l})-\min(\boldsymbol{P}^{l}))/(2^B-1)$ and $b^l=\min(\boldsymbol{P}^{l})$ are the scaling and zero factors, respectively; $\lfloor \cdot \rceil$ denotes the integer rounding operation. In this way, all elements in $\Tilde{\boldsymbol{P}}$ are mapped to the set $\{ 0, 1, \cdots, 2^B-1\}$ and thus stored as $B$-bits integers. We present the full algorithm steps of TWA in Algorithm~\ref{alg:TWA-algorithm}.

For memory savings, we only need to store the projection matrices $\boldsymbol{P}$ as $B$-bits integers, along with the compression factors $a^l$ and $b^l$, which are negligible. Compared to the original 32-bit format, we achieve a compression rate of ${32}/{B}$.
Note that the quantization technique can be incorporated into the distributed training scheme to further reduce the memory burden.
A detailed analysis of the memory requirements for the projection matrix $\boldsymbol{P}$ is presented in Table~\ref{tab:memory_projection}.
In this paper, we default $B=4$ for TWA, and an ablation of the quantization performance is presented in Section~\ref{sec:bits}.

\begin{algorithm}[t]
\small 
\caption{TWA Training} 
\label{alg:TWA-algorithm} 
\begin{algorithmic}[1] 
\Require  
    Sampled weights $ \{ \boldsymbol{w}_i \}_{i=1}^n$, 
    Batch size $b$, 
    Loss function $\mathcal{L}: \mathcal{W} \times \mathcal{X} \times \mathcal{Y} \rightarrow \mathbb{R}_+$, 
    Learning rate $\eta$, 
    Number of layers $L$, Quantization bits $B$, Regularization coefficients $\lambda$
\Ensure 
    Model trained with TWA
\State ${\boldsymbol{\overline w}} = \frac{1}{n} \sum_{i=1}^n \boldsymbol{w}_i$;
\State Extract layer-wise base vectors $\{\boldsymbol{e}^l_i\}_{i=1}^n$ with Eqn.~(\ref{equ:normalization});
\State Initialize $\boldsymbol{\beta}^{l,(0)} = \boldsymbol{0}$, $\boldsymbol{P}^l = [\boldsymbol{e}^l_1, \boldsymbol{e}^l_2, \cdots, \boldsymbol{e}^l_n]$ for $l\in\{1,2,\cdots,L\}$;
\State Quantize projection matrix $\boldsymbol{P}^l$ into $\Tilde{\boldsymbol{P}}^l$ with Eqn.~(\ref{equ:quantization});
\State $t = 0$;
\State  $\boldsymbol{w}_{\text{twa}}^{(t)}={\boldsymbol{\overline w}}$;
\While{\emph{not converged}}
    \State Sample batch data: $\mathcal{B} = \left \{ \left (\boldsymbol{x}_k, \boldsymbol{y}_k  \right )  
    |~\text{for}~k~\text{in the selected batch} \right \}$;
    \State Compute gradient: $\boldsymbol{g} = \nabla_{\boldsymbol{w}} \mathcal{L}_{\mathcal{B}}(\boldsymbol{w}_{\mathrm{twa}}^{(t)})$;
    \For{  $l \in \{1,2,\cdots,L\}$  }
    \State Update variables: 
    $
\boldsymbol{\beta}^{l,(t+1)} = (1-\eta\lambda)\boldsymbol{\beta}^{l,(t)} - \eta ( \tilde{\boldsymbol{P}}^l )^{\top} \boldsymbol{g}^l
    $;
    \State Update weights: 
    $
    \boldsymbol{w}^{l,(t+1)}_{\text{twa}} = {\boldsymbol{\overline w}}^{l} + {{\Tilde{\boldsymbol{P}}}^l} \boldsymbol{\beta}^{l,(t+1)}
    $;
\EndFor
    \State $t = t + 1$;
\EndWhile
\State \Return $\boldsymbol{w}_{\text{twa}}^{t}$
\end{algorithmic}
\end{algorithm}

\section{Numerical Experiments}
\label{sec:experiments}
In this section, we conduct numerical experiments on various computer vision and natural language processing tasks to demonstrate the efficiency and effectiveness of our proposed TWA approach.
First, we show that TWA can significantly accelerate the training of DNNs by averaging historical solutions from the head stage of training, especially for transformer-based architectures.
Next, we demonstrate that TWA significantly improves the performance of fine-tuned models across both single and multiple training configurations. Finally, we present ablation studies to further analyze the properties of TWA training.

\subsection{Accelerating Neural Network Training}
We first apply TWA to the head stage of training to accelerate the training of DNNs. In the head training stage, model weights are fast-evolving, and equal averaging like SWA may not be  enough and usually fails. 
Since TWA could adaptively adjust the weighting coefficients and reduce the estimation variance, it can be expected to work well in this stage and yield better performance. 
If so, it is promising to simultaneously attain generalization improvements and training efficiency. 
We conduct experiments over two representative computer vision and neural language processing tasks, i.e., image classification and machine translation, to evaluate the efficiency and effectiveness of the TWA scheme.

\subsubsection{Image Classification}
\label{sec:img-classification}
\paragraph{Setting.}
We experiment over two benchmark image classification datasets, i.e., CIFAR-100~\citep{krizhevsky2009learning} and ImageNet~\citep{deng2009imagenet}. Following~\cite{izmailov2018averaging, yang2019swalp}, we apply standard data preprocessing for experiments on CIFAR datasets and adopt the preprocessing and data augmentation procedures in the public Pytorch example on  ImageNet~\citep{paszke2017automatic}.
We use three representative architectures, VGG-16~\citep{simonyan2014very}, ResNet-18~\citep{he2016deep} and ViT-S/4~\citep{dosovitskiy2021an} on CIFAR experiments. For ImageNet, we use ResNet-18/50 \citep{he2016deep}, ViT-S/32, ViT-B/16~\citep{dosovitskiy2021an}.

For CIFAR training, 
we adopt a standard training protocol with a step-wise learning rate schedule. 
We run all experiments with three random seeds and report the mean test accuracy and standard deviation.
We use SGD optimizer with momentum $0.9$, weight decay $10^{-4}$, and batch size $128$. We train the models for $200$ epochs with an initial learning rate $0.1$ and decay it by $10$ at
the 100th and the 150th epochs. 
For ViT-S/4, we use AdamW as the base optimizer and train for 200 epochs with an initial learning rate of 0.001, weight decay of 0.1, and a cosine learning rate schedule.
For ImageNet training, we follow official PyTorch implementation\footnote{The implementation is available at: \url{https://github.com/pytorch/examples/tree/main/imagenet}.}.
We randomly split out 10\% and 2\% of training data for CIFAR-100 and ImageNet, respectively, as validation sets.
The validation data is used to select the best-performing model for base training and TWA and serves as supervision for optimizing the weighting coefficients in TWA-v.
For TWA, we sample solutions once after each epoch training for CIFAR  and ImageNet.
We apply the base optimizer for TWA training.
 We list the detailed training settings for the CIFAR experiments as follows. More training details can be found in Appendix~\ref{sec:training_details}.
\begin{itemize}
    \item Train the model for 200 epochs and use the validation set to select the best model.
    \item SWA: Average the model checkpoints from the first 100 epochs.
    \item TWA-t: Apply TWA to model solutions from the first 100 epochs, optimize coefficients using training data for 10 epochs, and select the optimal model based on validation performance. 
    \item TWA-v: Apply TWA to model solutions from the first 100 epochs, optimize using validation data for 10 epochs, and utilize the final model directly. 
\end{itemize}
For the ImageNet experiments, we maintain similar training settings while varying the training epochs. For instance, the notation ``$60+2$'' denotes that TWA is applied to model solutions collected from the first 60 epochs, followed by 2 additional epochs to optimize the coefficients using training data.

\begin{table}[!t]
  \caption{Top-1 accuracy (\%) and generalization gap on CIFAR-100 for head training with VGG / ResNet / ViT architectures. $^\dagger$ denotes epochs on the validation set. ``Gap'' denotes the generalization gap between training and test accuracies.}
\setlength\tabcolsep{3.4pt}
  \label{table-cifar}
  \centering
  {
  \setlength{\tabcolsep}{2 pt}
    \resizebox{\linewidth}{!}{
    \renewcommand{\arraystretch}{1.2} 
  \begin{tabular}{lccccccc}
    \toprule
    &\multicolumn{2}{c}{{ Base} ({$200$} {epochs})} 
    &{SWA~($100$ epochs)}
    &\multicolumn{2}{c}{{TWA-t} ($100+10${ epochs})} &\multicolumn{2}{c}{TWA-v ($100+10^\dagger$ {epochs})}\\
    \cmidrule(r){2-3} \cmidrule(r){4-4} \cmidrule(r){5-6} \cmidrule(r){7-8}
    {Model} &{Accuracy} &{Gap}  &{Accuracy}  &{Accuracy} &\multicolumn{1}{c}{{Gap}} &{Accuracy} &\multicolumn{1}{c}{{Gap}} \\
    \midrule
    {VGG-16} 
    &$71.99 \pm 0.17$ &$27.58$ &$69.58 \pm 0.24$ &$73.59 \pm 0.24$ &21.63 {\scriptsize \color{teal} ($\downarrow5.59$)} &$73 .80\pm 0.24$ &$18.49$ {\scriptsize \color{teal} ($\downarrow9.09$)}
    \\
    {ResNet-18} &$73.45 \pm 0.22$ &$25.99$ &$73.02 \pm 0.55$ &$74.03 \pm 0.44$ &$25.44$ {\scriptsize \color{teal} ($\downarrow0.50$)} &$74.78 \pm 0.33$ &$24.76$ {\scriptsize \color{teal} ($\downarrow1.23$)}\\
    {ViT-S/4}  &$58.86 \pm 0.28$ &41.10 &$39.66\pm 0.44$ &$58.64 \pm 0.23$ &41.05 {\scriptsize \color{teal} ($\downarrow0.05$)} &${60.57} \pm 0.28$ &30.13 {\scriptsize \color{teal} ($\downarrow9.97$)}\\
    \bottomrule 
  \end{tabular}
  }
  }
\end{table}

\noindent
\paragraph{Results.}
We first investigate the experiments on CIFAR datasets. The base training schedule contains 200 epochs and we take the first 100 epochs' explorations for TWA. 
The results are given in Table~\ref{table-cifar}. It can be observed that TWA achieves comparable or even better performance compared to regular SGD training with a significant reduction in the generalization gap. 
For instance, TWA-v attains $1.81\%$ accuracy improvement on CIFAR-100 with VGG-16, while the generalization gap is reduced by $9.09\%$. 
This suggests that a better solution could already be composed by weighted averaging these historical solutions without further training at finer learning rates, which may otherwise lead to overfitting problems and harm generalization.
In comparison, we also apply SWA to average these samples, which shows degraded performance due to the existence of estimation error. Compared to TWA-t, TWA-v achieves much better performance in both test accuracy and generalization gap, confirming the effectiveness of using the validation set to optimize the coefficients. Notably, such improvement is more obvious on ViT-S/4, where TWA-v outperforms TWA-t by 1.93\%.

For ImageNet, the effort required for training each epoch is significantly greater, making efficient methods to reduce the number of training epochs highly desirable. The comparison results of SGD/SWA/TWA are presented in Table \ref{table-imagenet}.
Beyond narrowing the generalization gap between training and test data, TWA achieves performance comparable to, or even better than, standard SGD training with 90 epochs by averaging the historical solutions from the first 60 epochs.
For comparison, Lookahead~\citep{zhang2019lookahead} is another advanced optimizer recently proposed for improving convergence and reported $75.49\%$ accuracy at the 60th epoch (Table 2 in \cite{zhang2019lookahead}) with an aggressive learning rate decay (i.e., the learning rate is decayed at the 30th, 48th, and 58th epochs). In contrast, our TWA-v method achieves $75.90\%$ accuracy with the same budget, simply employing conventional decay.
Again, This improvement is particularly pronounced with ViT architectures, where TWA-v can achieve a 3.23\% accuracy increase with ViT-B/16 and a 2.32\% increase with ViT-S/32 compared to standard AdamW training while requiring over 30\% fewer training epochs.
In fact, TWA-v, with approximately 60 epochs of training, outperforms AdamW training based on a 300-epoch baseline (e.g., 74.6\% with ViT-B/16) as reported in \cite{chen2022when}. This shows the great potential of TWA-v for accelerating the training of transformer-based architectures.
\begin{table}[!t]
  \caption{Top-1 accuracy (\%) and generalization gap (\%) on ImageNet for head training with ResNet and ViT architectures. $^\dagger$ denotes epochs on the validation set. ``Gap'' refers to the generalization gap, the difference between training and test accuracies.}
  \setlength{\tabcolsep}{1.5 pt}
  \label{table-imagenet}
  \centering
    \resizebox{\linewidth}{!}{
  {\footnotesize
    \renewcommand{\arraystretch}{1.2} 
  \begin{tabular}{lccccrcr}
    \toprule
    &\multicolumn{2}{c}{{Base ($90$ epochs)}} &\multicolumn{1}{c}{SWA}($60$ {epochs})  &\multicolumn{2}{c}{{TWA-t ($60+2$ epochs)}} &\multicolumn{2}{c}{{TWA-v ($60+5^\dagger$ epochs)}} \\
    \cmidrule(r){2-3} \cmidrule(r){4-4} \cmidrule(r){5-6} \cmidrule(r){7-8}
   {Model} &{Accuracy} &{Gap}  &{Accuracy} &{Accuracy}  &\multicolumn{1}{c}{Gap } &{Accuracy}  &\multicolumn{1}{c}{Gap } \\
    \midrule
    {ResNet-18 } &69.53 &-0.45 &64.03 &69.42 &$-0.76$ {\scriptsize \color{teal} ($\downarrow0.31$)} &69.66 &$-1.13$ {\scriptsize \color{teal} ($\downarrow0.68$)}\\
    {ResNet-50} &$75.82$ &$0.25$ &67.66 &$75.90$ &$-0.68$ {\scriptsize \color{teal} ($\downarrow0.93$)} &$76.11$ &$-0.99$ {\scriptsize \color{teal} ($\downarrow1.24$)}\\
    {ViT-S/32} &66.95 &18.64 &62.45  &67.39 &10.62 {\scriptsize \color{teal} ($\downarrow8.02$)} &{69.27} &6.31 {\scriptsize \color{teal} ($\downarrow12.33$)}\\
    {ViT-B/16} &72.79 &20.07 &67.12  &73.20 &14.21 {\scriptsize \color{teal} ($\downarrow5.86$)} &{76.02} &7.67 {\scriptsize \color{teal} ($\downarrow12.40$)}\\
    
    \bottomrule
  \end{tabular}
  }
  }
\end{table}

\paragraph{Comparison of training time.}
In Table~\ref{tab:cost}, we compare the training time cost of different methods. For TWA training, the training time is comprised of two components: Stage 1 training for collecting historical training and Stage 2 training for TWA training.
Apart from the good performance in accuracy, TWA-v achieves around 45\% savings in total training time on CIFAR-100 and 32\% on ImageNet. 
Compared to TWA-t, TWA-v can further reduce the Stage 2 training time by optimizing on the much smaller validation set, while enabling weight averaging with much better generalization performance.

\begin{table}[htbp]
    \caption{Comparison of total training time cost.}
    \vspace{2mm}
    \label{tab:cost}
    \centering
    \small
    \renewcommand{\arraystretch}{1.2} 
    \resizebox{\linewidth}{!}{
    \begin{tabular}{lccccccr}
    \toprule
         Datasets &Model &Method &Test Accuracy (\%) &Stage 1 &Stage 2 &Total &Base\% \\
    \midrule
         CIFAR-100 &ResNet-18 &Base (SGD) &73.45 &0.51h &- &0.51h &100\%\\
         & &TWA-t &74.03 &0.26h &0.09h &0.35h &69\%\\
         & &TWA-v &74.78 &0.26h &0.02h &0.28h &55\%\\
    \midrule
         ImageNet &ViT-B/16 &Base (AdamW) &72.79 &51.8h &-&51.8h &100\%\\
         & &TWA-t &73.08 &34.5h &3.0h &37.5h &72\%\\
         & &TWA-v &76.02 &34.5h &0.5h &35.0h &68\%\\
    \bottomrule
    \end{tabular}}
\end{table}

\begin{figure}[!t]
    \centering
    \includegraphics[width=0.6\linewidth]{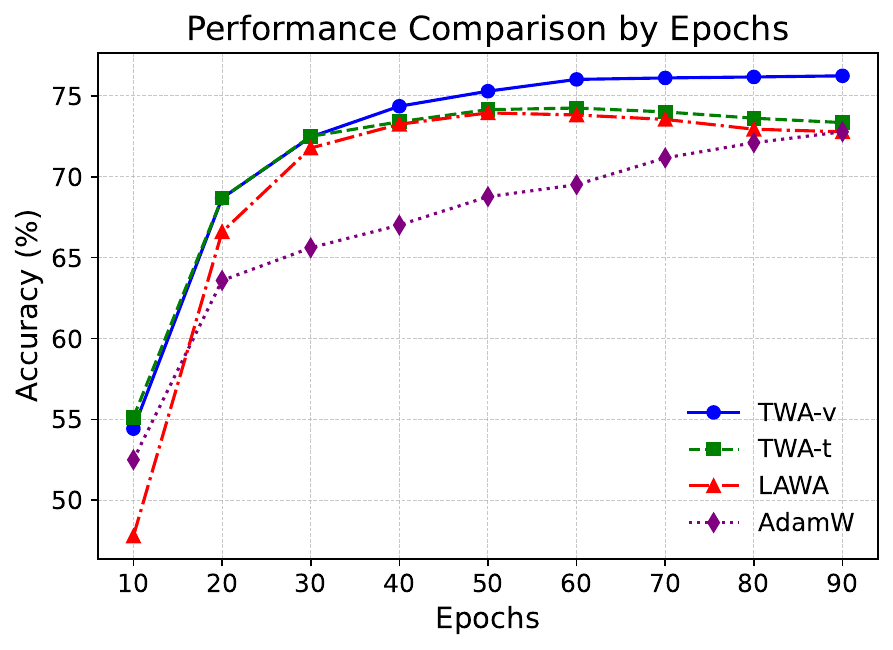}
    \caption{Performance comparison with LAWA across different averaging epochs. The experiments are conducted on ImageNet using ViT-B/16.}
    \label{fig:lawa}
\end{figure}

\paragraph{Comparison with LAWA.}
LAWA~\citep{kaddour2022stop} is a recently proposed method for accelerating the training of DNNs by averaging the latest weights checkpoints. We compare LAWA with TWA across different averaging epochs. 
We sample model weights once after each epoch of training for LAWA, maintaining consistency with TWA.
We set the horizon $t=5$ for ImageNet as suggested by \cite{kaddour2022stop}, and the accuracy reached by AdamW before averaging is also given for reference. 
The results are presented in Figure~\ref{fig:lawa}.

Generally, utilizing more epochs of explorations can provide a better estimation for the final minimum, and it could be observed that the model's performance is consistently improved with more epochs of explorations.
Notably, although each historical solution in a relatively short period of explorations by AdamW may not be good, satisfactory solutions have already emerged in the subspace spanned by these solutions. 
Through proper optimization within this subspace, TWA can identify them.
For instance, on ImageNet with the ViT-B/16 model, averaging over 30 epochs via TWA matches the final performance of regular AdamW training. Moreover, TWA consistently outperforms LAWA across different averaging epochs, and the advantage of TWA-v over both TWA-t and LAWA becomes more pronounced as the training epochs increase. Notably, in the later stages of training, the performance of TWA-t and LAWA tends to decline, while TWA-v continues to improve, highlighting the effectiveness of validation loss supervision.

\paragraph{Comparison with training without splitting validation set.} 
One might be concerned that splitting out a validation set could reduce the available training data and thus harm the performance of base training. To investigate this, we compare with training without a validation set, i.e., using 100\% of the training data for model training and evaluating with the last model. The results are shown in Table \ref{tab:no_split}. We observe that indeed splitting out a small validation set slightly degrades the performance of base training compared to training without splitting a validation set. However, such degradation can be made up by applying TWA-v to average over the entire training trajectory---and can even achieve better results, e.g. by +0.50\% with ResNet-18 and +3.14\% on ViT-B/16. This provides a new perspective on utilizing available training data for DNNs' training: \emph{instead of using the entire dataset to train a single model, we can allocate a portion of the training data for ensembling historical solutions to achieve better performance.}

\begin{table}[!t]
    \caption{Ablation study on the usage of a held-out validation set for ImageNet. While separating a small validation set (2\% of the training set) may slightly reduce the base training performance, it enables us to achieve better overall results by effectively ensembling historical solutions using TWA-v with the validation data.}
    \label{tab:no_split}
    \centering
    \renewcommand{\arraystretch}{1.2} 
    \small
    \begin{tabular}{lccc}
    \toprule
         Model &Method &Validation Ratio~(\%) &Test Accuracy~(\%)  \\
    \midrule
    ResNet-18 &Base (SGD) &w/o splitting &69.82 \\
    &Base (SGD) &2 &69.53\\
    &TWA-v &2 &70.32\\
    \midrule
        ViT-B/16 &Base (AdamW) &w/o splitting &73.0\\
         &Base (AdamW) &2 &72.79\\
         &TWA-v &2 &76.14 \\
    \bottomrule
    \end{tabular}
\end{table}

\subsubsection{Machine Translation}
\label{sec:machine_translation}
\paragraph{Settings.}
In this study, we train a transformer based  model~\citep{vaswani2017attention} to perform English-to-German translation on WMT2014 dataset~\citep{bojar2014findings}.
The size of the embedding layer is set to 512.
Similar to \cite{vaswani2017attention},
we utilize byte-pair encoding and construct a common vocabulary of 32,000 tokens.
We use Adam optimizer with a weight decay of 0.0001, an initial learning rate of 0.0005, a batch size of 256, and a dropout rate of 0.1. 
We train the model for 100k steps using a ReduceLROnPlateau schedule following \cite{bisla2022low}. For TWA, we sample the weights once per 1,000 steps and take for a total of the first 50 model checkpoints (corresponding to the initial 50k steps of training) to achieve training efficiency. We then train for a total 1000 steps with Adam optimizer and a constant learning rate 0.01. For TWA-v, the validation data are utilized for optimizing the coefficients, whereas for other methods, the best-performing model is selected based on the validation loss.

\noindent
\paragraph{Results.}
In Table~\ref{tab:MT}, we report the BLEU scores of test data for different methods. We observe that applying TWA to the historical solutions from the first 50k iterations has already surpassed the performance of regular Adam training with 100k steps. 
This will lead to a significant time/computation saving of around 50\% compared to regular training. 
Moreover, TWA-v can further improve the test BLUE score by 0.18 over TWA-t.
We report the generalization gap between the training and test data on the BLUE score metric as well, and the results confirm the effectiveness of TWA in improving generalization. 


\begin{table}[htbp]
    \centering
    \caption{BLUE scores for English to German translation on WMT2014 datasets. The generalization gap for Adam 50k and SWA is omitted (-) due to their subpar performance compared to the final results.}
    \vspace{2mm}
    \label{tab:MT}
    \renewcommand{\arraystretch}{1.2} 
    \small
    \begin{tabular}{lrcc}
    \toprule
         {Optimizer} &\multicolumn{1}{c}{Steps} 
         &{Test BLUE Scores} &{Gap} \\
         \midrule
         {Adam } &{100k}(100\%)  &$20.54 \pm 0.18$ &$0.35$ \\
         {Adam } &{50k}~(50\%)  &$18.95\pm0.31$ & -\\
         {SWA } &{50k}~(50\%)  &$16.84\pm0.22$ &-\\
          {TWA-t} &{50k+1k}~(51\%) &$20.62\pm0.19$ &$0.31$ {\scriptsize \color{teal} ($\downarrow0.04$)}  \\
         {TWA-v } &{50k+1k}~(51\%) &$20.80\pm0.22$ &$0.29$ {\scriptsize \color{teal} ($\downarrow0.06$)} \\
    \bottomrule
    \end{tabular}
\end{table}


\subsection{Better Fine-tuning Performance}
\label{sec:finetune}
Next, we turn our attention to fine-tuning tasks. In our previous experiments, we focused on a single training run and aimed to achieve training efficiency by sufficiently utilizing the historical solutions generated during the training process. In this section, we seek to improve the model performance by leveraging the flexibility and good generalization ability of our trainable weight averaging scheme.
Specifically, we aim to enhance model performance by fully utilizing the solutions from single or multiple fine-tuning configurations.
Previously, greedy soup~\citep{wortsman2022model} demonstrated state-of-the-art performance by adding weights to the weight-averaging ``soup'' in a greedy manner.
However, TWA has the potential to provide even better performance by leveraging trainable weighting coefficients, rather than relying on a fixed averaging strategy. 

For fine-tuning tasks, the models to be averaged have generally adapted well to training data, as evidenced by their low training loss. Therefore, training them would provide little meaningful supervision. Motivated by  \cite{wortsman2022model}, we directly train over the held-out validation set. This makes sense because the validation data are ``new'' to these models, and
subspace training involves a small number of independent variables, so a small validation set is expected to be sufficient to well train them.
In this section, we conduct experiments on image classification and language modeling tasks to demonstrate the superiority of our scheme.

\subsubsection{Image Classification}
\label{sec:finetune-image}
\noindent
\paragraph{Setting.}
We conduct experiments on CIFAR-10~\citep{krizhevsky2009learning} and ImageNet~\citep{deng2009imagenet} datasets using pretrained CLIP ViT-B/32 model~\citep{radford2021learning}. We utilize the publicly available fine-tuned checkpoints from \cite{wortsman2022model}, which include 5 models for CIFAR-10 and 72 models for ImageNet\footnote{The model checkpoints are available at \url{https://github.com/mlfoundations/model-soups}.}. These models are obtained by a random hyperparameter search over learning rate, weight decay, training epochs, label smoothing, and data augmentation.
For TWA, we train the models using  AdamW optimizer with a learning rate of 0.01 and a cosine learning rate schedule.
For TWA, we train for 2 epochs on the training datasets, selecting the best model based on the held-out validation set, and for TWA-v, we directly train for 5 epochs on the validation set and report the final performance on the test set.

\paragraph{Results.}
We present the results for CIFAR-10 and ImageNet datasets in Table~\ref{tab:finetune-cifar} and Table~\ref{tab:finetune-img}, respectively. We compare our TWA with SWA~\citep{izmailov2018averaging} and greedy soup~\citep{wortsman2022model}, and also list the performance of the best and second-best individual models for reference.
For CIFAR-10, we observe that both SWA and greedy soup can significantly improve performance compared to the best individual model, confirming the effectiveness of weight averaging. However, our TWA can further improve the performance of greedy soup: TWA achieves an additional $+0.05\%$ improvement, while TWA-v provides a further $+0.10\%$ gain. Note that for CIFAR-10, the performance is already very high, making further improvements challenging. Our enhancements represent $83\%$ and $167\%$ increases over the performance gains of greedy soup relative to SWA.

\begin{table}[htbp]
    \centering
    \caption{Results on CIFAR-10 for fine-tuning CLIP ViT-B/32 with multiple configurations.}
    \vspace{2mm}
    \small
    \label{tab:finetune-cifar}
    \renewcommand{\arraystretch}{1.2} 
    \begin{tabular}{lc}
    \toprule
         Method &Test Accuracy (\%)  \\
         \midrule
         Best individual model&$98.03$ \\
         Second best individual model&$97.83$ \\
         \midrule
         SWA &$98.39$\\
         Greedy soup &$98.45$\\
         TWA-t &$98.50$\\
         TWA-v &${98.55}$\\
    \bottomrule
    \end{tabular}
\end{table}
We then focus on the results of ImageNet datasets in Table~\ref{tab:finetune-img}, which involves a total of 72 fine-tuned models. Equal averaging, as done with SWA,  does not yield any performance improvement compared to the best individual model and, in fact, results in a slight degradation. 
This occurs because the model configurations are diverse, and equal averaging can be significantly influenced by poorer solutions, leading to estimation errors. 
Greedy soup effectively selects an optimal subset of weights for averaging, providing an accuracy gain of 0.65\% over the best individual model.
Our TWA-v delivers further significant improvements, i.e., $+0.58\%$ over greedy soup, respectively, confirming the superiority of optimizing the weighting coefficients. We can also observe that TWA-v offers greater advantages over TWA-t, as models in fine-tuning scenarios are often overfitted, and the training loss provides only minimal supervision.
\begin{table}[htbp]
    \centering
    \caption{Results on ImageNet for fine-tuning CLIP ViT-B/32 with multiple configurations.}
    \vspace{2mm}
    \small
    \label{tab:finetune-img}
    \renewcommand{\arraystretch}{1.2} 
    \begin{tabular}{lc}
    \toprule
         Method &Test Accuracy (\%)  \\
         \midrule
         Best individual model&$80.38$ \\
         Second best individual model&$79.89$ \\
         \midrule
         SWA &$79.97$\\
         Greedy soup &$81.03$\\
         TWA-t &$80.56$\\
         TWA-v &${81.61}$\\
    \bottomrule
    \end{tabular}
\end{table}


\subsubsection{Language Modeling}
\label{sec:language_modeling}
\paragraph{Setting.}
TWA can enhance fine-tuning performance not only by utilizing solutions fine-tuned with multiple configurations but also by leveraging solutions from a single training trajectory. In this experiment, we fine-tune a pre-trained language model GPT-2~\citep{radford2019language} for causal language modeling tasks. 
Specifically, we use the raw WikiText-2 datasets~\citep{merity2016pointer} (no tokens are replaced before tokenization) following the official HuggingFace Implementation\footnote{The implementation is available at \url{https://github.com/huggingface/transformers/tree/main/examples/pytorch/language-modeling}.}. The model is trained using AdamW optimizer for  1,000 steps, with a learning rate of 5e-5, batch size of 8, and a linear learning rate schedule. 
During the training, we sample the weights once per 100 steps for weight averaging, resulting in 10 model checkpoints. For TWA, we train for 100 steps using AdamW optimizer with a learning rate of 0.01 and a weight decay of 0.01.

\paragraph{Results.}
We report the perplexity score of different methods in Table \ref{tab:wiki}. For comparison, we consider the best/second best individual models among the 10 collected weights, as well as SWA and greedy soup solutions. 
We observe that again, TWA-v significantly improves the performance of SWA and greedy soup by 0.30 and 0.19 in perplexity, respectively. 
It is worth noting that the fine-tuned models in this experiment are derived from a single training run, which limits the performance gains compared to Section \ref{sec:finetune-image}. Nevertheless, these results confirm that TWA can be effectively applied to fine-tuning tasks, whether under single or multiple training configurations.

\begin{table}[htbp]
    \centering
    \caption{Results on Wikitext-2 for fine-tuning GPT-2 with a single configuration.}
    \renewcommand{\arraystretch}{1.2} 
    \vspace{2mm}
    \small
    \label{tab:wiki}
    \begin{tabular}{lc}
    \toprule
    Method &Perplexity ($\downarrow$) \\
    \midrule
         Best individual model &$20.50$  \\
         Second best individual model &$20.51$ \\
         \midrule
         SWA &$20.55$\\
         Greedy soup &$20.44$ \\
         TWA-t &$20.42$ \\
         TWA-v &${20.25}$ \\
    \bottomrule 
    \end{tabular}
\end{table}


\subsection{Ablation Study}
\label{sec:ablation}

\subsubsection{Subspace Construction}
\label{sec:subspace-extraction}
We compare the execution time of different subspace construction methods in Table~\ref{tab:extraction}. Specifically, we experiment with the ResNet-50 model on ImageNet and a total of 60 historical solutions as specified in Section~\ref{sec:img-classification}.  We configure the number of GPUs to four. Our results show that compared with the orthogonalization approach in the previous conference version~\citep{li2023trainable}, the decentralization and normalization approach dramatically reduces the extraction time by orders of magnitude, resulting in significant savings in floating-point numerical operations. This improvement is due to the orthogonalization approach being a sequential process—obtaining orthogonal vectors one by one—while our decentralization and normalization method can be efficiently executed in parallel.
\begin{table}[!t]
    \centering
    \caption{Execution time of different subspace construction methods.}
    \vspace{2mm}
    \small
    \label{tab:extraction}
    \renewcommand{\arraystretch}{1.2} 
    \begin{tabular}{lc}
    \toprule
         Extraction Method &{Time (seconds)} \\
         \midrule
         Orthogonalization~\citep{li2023trainable} &382.6  \\
         Decentralization and normalization~(ours) &6.3   \\
    \bottomrule
    \end{tabular}
\end{table}

\begin{table}[htbp]
    \centering
    \footnotesize
    \caption{Averaged epoch training time and memory burden comparisons of SGD and TWA with DDP training. The experiments are conducted on ImageNet with ResNet-50 and measured on NVIDIA Tesla A100 40G
GPUs. }
    \label{tab:ddp}
    \renewcommand{\arraystretch}{1.3} 
    \setlength{\tabcolsep}{8pt} 
    \begin{tabular}{cccc}
        \toprule
        & \multicolumn{3}{c}{Time (seconds)} \\ 
        \cmidrule(lr){2-4} 
        \#GPUs & SGD & TWA (w/o quantization) & TWA \\ 
        \midrule
        1 & 1638 & 1692 (+3.3\%) & 1725 (+5.3\%) \\
        2 & 824 & 862 (+4.6\%) & 865 (+5.0\%) \\
        4 & 420 & 432 (+2.8\%) & 435 (+3.5\%) \\ 
        \midrule
        & \multicolumn{3}{c}{Memory (MB)} \\ 
        \cmidrule(lr){2-4} 
        \#GPUs & SGD & TWA (w/o quantization) & TWA \\ 
        \midrule
        1 & 20287 & 26287 (+29.6\%) & 21037 (+3.7\%) \\
        2 & 20383 & 23383 (+14.7\%) & 20758 (+1.9\%) \\
        4 & 20875 & 22375 (+7.2\%) & 21062 (+1.0\%) \\ 
        \bottomrule
    \end{tabular}
\end{table} 

\subsubsection{Training Speed and Memory}
\label{sec:training-speed}
We numerically measure the averaged epoch training time and memory burden for SGD and TWA under the DDP training setting.
Specifically, we experiment with the ResNet-50 model on ImageNet, using 1, 2, and 4 GPUs with a batch size of 256 per GPU, and utilize a total of 60 historical solutions as specified in Section~\ref{sec:img-classification}. 
For TWA, we select the TWA-t version that optimizes over the training set, keeping the same as SGD.
The experiments are conducted on NVIDIA Tesla A100 40G GPUs. 
From the results reported in Table~\ref{tab:ddp},
we observe that TWA introduces minimal additional costs, e.g., +5.3\% on time cost and +3.7\% on memory burden with one GPU, compared with regular SGD training. 
With more GPUs, the additional memory burden can be further reduced to +1.0\%, which shows the effectiveness of our distributed training scheme.
Then compared to TWA without quantization on the projection matrix, we observe that the additional memory overhead is effectively reduced by around 8x with our 4-bit quantization scheme, though it slightly increases the training time (e.g., +2\%). In fact, the gradient projection in our subspace training scheme incurs negligible extra training time compared to regular SGD, as matrix multiplication operations are highly efficient on GPUs. Overall, these results demonstrate that TWA provides an efficient and scalable weighted averaging approach for large-scale problems.


\subsubsection{Effects of Layer-wise Processing}
\label{sec:layer}
In this subsection, we study the effects of layer-wise processing. 
Table~\ref{tab:layer-wise} compares the performance of TWA-v with and without layer-wise processing and projection matrix quantization. 
The results show that layer-wise processing significantly improves test performance—by 0.53\% without quantization and more substantially by 2.08\% with quantization.
These findings indicate that layer-wise processing enables more effective quantization of the projection matrix without compromising performance, likely due to the varying magnitude of values across different layers.
\begin{table}[htbp]
    \caption{Ablation study of layer-wise processing and projection matrix quantization. The experiments are conducted on ImageNet with CLIP ViT-B/32.}
    \small
    \label{tab:layer-wise}
    \centering
    \renewcommand{\arraystretch}{1.2} 
    \begin{tabular}{lc}
    \toprule
         Method  &Test Accuracy (\%)  \\
    \midrule
         TWA-v &81.61 \\
         TWA-v+w/o quantization &81.67 \\
         TWA-v+w/o layer-wise  &79.53\\
         TWA-v+w/o layer-wise+w/o quantization  &81.14\\
    \bottomrule
    \end{tabular}
\end{table}

\subsubsection{Effects of Weight Decoupling}
\label{sec:decouple}
We then evaluate the performance of different subspace construction methods. From the results in Table~\ref{tab:subspace-performance}, we observe that without applying any orthogonalization methods to the weights $\{\boldsymbol{w}_i\}_{i=1}^n$, the cosine similarity between them remains very high (e.g., 99\%), which degrades training performance. While orthogonalization effectively reduces correlations between the bases, its performance is suboptimal, possibly due to the numerical errors introduced by the sequential orthogonalization process. In contrast, our proposed decentralization and normalization approach effectively reduces the correlations between the bases while introducing fewer numerical operations, achieving the best performance.

\begin{table}[htbp]
    \caption{Performance of different subspace construction methods. ``Correlations'' refers to the average cosine similarity between pairs of the obtained bases. The experiments are conducted on ImageNet with CLIP ViT-B/32.}
    \small
    \label{tab:subspace-performance}
    \centering
    \renewcommand{\arraystretch}{1.2} 
    \begin{tabular}{lcc}
    \toprule
         Method & Correlations &Test Accuracy (\%)  \\
    \midrule
         W/o orthogonalization &0.99 &81.02\\
         Orthogonalization &0 &81.35\\
         Decentralization and normalization &0.19 &81.61 \\
    \bottomrule
    \end{tabular}
\end{table}

\subsubsection{Effects of Fine-tuned Models' Number}
We vary the number of fine-tuned models in Table \ref{tab:finetune-img} and study its impacts on TWA performance. The results are in Table \ref{tab:finetune-ablation}. As one could expect, the performance of all methods is consistently increasing with more fine-tuned models. However, TWA-v demonstrates significantly higher efficiency in utilizing fine-tuned models compared to competing methods. For example, TWA-v achieves comparable performance to greedy soup with only 18 models, while greedy soup requires 72 models. Furthermore, with 36 models, TWA-v surpasses its greedy soup counterpart with 72 models by a notable margin of 0.36\%. These results highlight that TWA-v can significantly reduce computational costs in fine-tuning various models—by up to 4x—while achieving comparable performance to the greedy soup.

\begin{table}[htbp]
    \centering
    \caption{Fine-tuning results on ImageNet with CLIP ViT-B/32.}
    \vspace{2mm}
    \small
    \label{tab:finetune-ablation}
    \renewcommand{\arraystretch}{1.2} 
    \begin{tabular}{lccc}
    \toprule
          &\multicolumn{3}{c}{Test Accuracy (\%)}  \\
          Method &N=18 &N=36 &N=72 \\
         \midrule
         Best individual model &$78.73$ &$79.76$ &$80.38$ \\
         SWA &$79.14$ &$79.55$ &$79.97$\\
         Greedy soup &$79.87$ &$80.86$  &$81.03$\\
         TWA-t &$79.81$ &$80.32$  &$80.56$\\
         TWA-v &{$80.96$} &{$81.42$} &${81.61}$\\
    \bottomrule
    \end{tabular}
\end{table}

\subsubsection{Effects of Quantization Bits}
\label{sec:bits}
\begin{figure}[htbp]
    \centering
    \begin{subfigure}{0.48\textwidth}
        \centering
        \includegraphics[width=\textwidth]{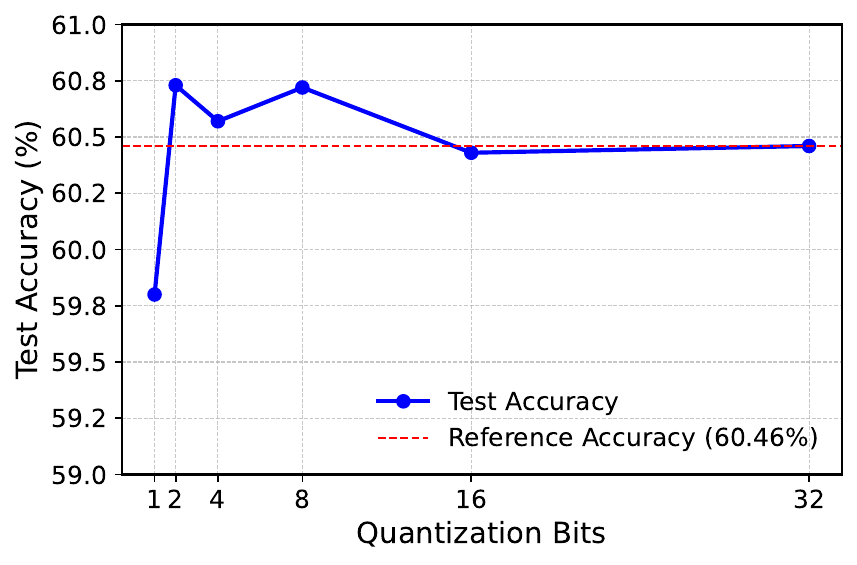}
        \caption{ViT-S/4 on CIFAR-100 for head training}
    \end{subfigure}\hfill
    \begin{subfigure}{0.48\textwidth}
        \centering
        \includegraphics[width=\textwidth]{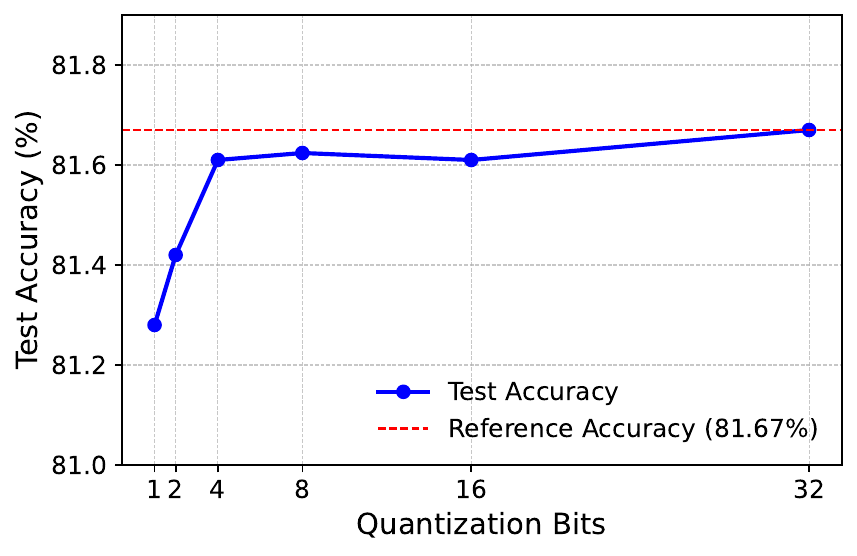}
        \caption{ViT-B/32 on ImageNet for fine-tuning}
    \end{subfigure}
    \caption{Performance of TWA-v with different quantization bits for projection matrix.}
    \label{fig:bits}
\end{figure}

In this section, we study the impact of quantization bits on the TWA-v performance. We experiment over two scenarios: accelerating training by averaging historical solutions, where we use CIFAR-100 with ViT-S/4, and better fine-tuning setting where we use ImageNet with CLIP ViT-B/32. The original projection matrix $\boldsymbol{P}$ is stored as \texttt{float32} defaulted as model parameters, and we quantize it to different bit levels among $\{1,2,3,8,16\}$. The results are shown in Figure~\ref{fig:bits}.
We observe that overall, TWA-v can achieve good performance even with 1-bit quantization, demonstrating the intriguing quantization properties of the projection matrix.
Moreover, with 4-bit quantization, TWA-v achieves performance comparable to full-precision training, which we adopt as the default setting for our usage.

\section{Conclusion}
\label{sec:conclusion}

In this work, we propose \emph{Trainable Weight Averaging} (TWA), an efficient framework that enables weight averaging with learnable weighting coefficients. It extends the manually defined weighting coefficients as in previous works to a trainable manner, which endows with much greater flexibility and enables handling weights from different stages and configurations.
We design an efficient parallel training scheme to cope with large-scale training and propose quantization schemes for the projection matrix to achieve memory efficiency.
Additionally, we derive two variants, TWA-t and TWA-v, based on the data used for training, and show that TWA-v allows for more efficient and effective averaging when validation data is available.
Extensive experiments on both efficient training and fine-tuning tasks demonstrate the effectiveness and efficiency of our approach.

\paragraph{Limitation and Future Works.} 
Although TWA can significantly enhance training efficiency and improve performance for DNN's training by effectively leveraging historical solutions, the selection of these solutions can influence overall performance, as illustrated in Figure~\ref{fig:lawa}. In practice, the specific strategy for selecting historical solutions can be predetermined in conjunction with the original training schedule for standard training.
Currently, TWA can only manage weights from a single linear model; however, it holds promise for extending algorithms to merge weights from different linear models or even from different architectures. 
Furthermore, despite the relatively small number of training variables, there remains a risk of overfitting, which can potentially degrade generalization performance.

There are many promising future directions for TWA, both in terms of practical applications and theoretical analysis. These include:
\textbf{1)} Combining TWA with other lightweight  methods, such as LoRA \citep{hu2022lora}, to facilitate more efficient adapter training and better fine-tuning performance;
\textbf{2)} Applying TWA to other precision-crucial scenarios, such as low precision network training \citep{yang2019swalp} and network quantization \citep{zhou2017incremental}; 
\textbf{3)} Understanding the relationship between the number of held-out samples required for subspace training and the dimension of the subspace, as well as its impact on generalization performance;
and \textbf{4)} Exploring advanced quantization schemes for compressing the projection matrix to enable more efficient TWA training.

\vskip 0.2in
\bibliography{main}

\begin{thebibliography}{47}
\providecommand{\natexlab}[1]{#1}
\providecommand{\url}[1]{\texttt{#1}}
\expandafter\ifx\csname urlstyle\endcsname\relax
  \providecommand{\doi}[1]{doi: #1}\else
  \providecommand{\doi}{doi: \begingroup \urlstyle{rm}\Url}\fi

\bibitem[Bisla et~al.(2022)Bisla, Wang, and Choromanska]{bisla2022low}
Devansh Bisla, Jing Wang, and Anna Choromanska.
\newblock Low-pass filtering sgd for recovering flat optima in the deep
  learning optimization landscape.
\newblock In \emph{International Conference on Artificial Intelligence and
  Statistics (AISTATIS)}, 2022.

\bibitem[Bojar et~al.(2014)Bojar, Buck, Federmann, Haddow, Koehn, Leveling,
  Monz, Pecina, Post, Saint-Amand, et~al.]{bojar2014findings}
Ond{\v{r}}ej Bojar, Christian Buck, Christian Federmann, Barry Haddow, Philipp
  Koehn, Johannes Leveling, Christof Monz, Pavel Pecina, Matt Post, Herve
  Saint-Amand, et~al.
\newblock Findings of the 2014 workshop on statistical machine translation.
\newblock In \emph{Proceedings of the Ninth Workshop on Statistical Machine
  Translation}, 2014.

\bibitem[Chen et~al.(2022)Chen, Hsieh, and Gong]{chen2022when}
Xiangning Chen, Cho-Jui Hsieh, and Boqing Gong.
\newblock When vision transformers outperform resnets without pre-training or
  strong data augmentations.
\newblock In \emph{International Conference on Learning Representations}, 2022.
\newblock URL \url{https://openreview.net/forum?id=LtKcMgGOeLt}.

\bibitem[Croce et~al.(2023)Croce, Rebuffi, Shelhamer, and
  Gowal]{croce2023seasoning}
Francesco Croce, Sylvestre-Alvise Rebuffi, Evan Shelhamer, and Sven Gowal.
\newblock Seasoning model soups for robustness to adversarial and natural
  distribution shifts.
\newblock In \emph{Proceedings of the IEEE Conference on Computer Vision and
  Pattern Recognition (CVPR)}, 2023.

\bibitem[Deng et~al.(2009)Deng, Dong, Socher, Li, Li, and
  Fei-Fei]{deng2009imagenet}
Jia Deng, Wei Dong, Richard Socher, Li-Jia Li, Kai Li, and Li~Fei-Fei.
\newblock Imagenet: A large-scale hierarchical image database.
\newblock In \emph{Proceedings of the IEEE Conference on Computer Vision and
  Pattern Recognition (CVPR)}, pages 248--255, 2009.

\bibitem[Dettmers et~al.(2022)Dettmers, Lewis, Belkada, and
  Zettlemoyer]{dettmers2022gpt3}
Tim Dettmers, Mike Lewis, Younes Belkada, and Luke Zettlemoyer.
\newblock Gpt3. int8 (): 8-bit matrix multiplication for transformers at scale.
\newblock \emph{Advances in Neural Information Processing Systems},
  35:\penalty0 30318--30332, 2022.

\bibitem[Dettmers et~al.(2024)Dettmers, Pagnoni, Holtzman, and
  Zettlemoyer]{dettmers2024qlora}
Tim Dettmers, Artidoro Pagnoni, Ari Holtzman, and Luke Zettlemoyer.
\newblock Qlora: Efficient finetuning of quantized llms.
\newblock \emph{Advances in Neural Information Processing Systems}, 36, 2024.

\bibitem[Dosovitskiy et~al.(2021)Dosovitskiy, Beyer, Kolesnikov, Weissenborn,
  Zhai, Unterthiner, Dehghani, Minderer, Heigold, Gelly, Uszkoreit, and
  Houlsby]{dosovitskiy2021an}
Alexey Dosovitskiy, Lucas Beyer, Alexander Kolesnikov, Dirk Weissenborn,
  Xiaohua Zhai, Thomas Unterthiner, Mostafa Dehghani, Matthias Minderer, Georg
  Heigold, Sylvain Gelly, Jakob Uszkoreit, and Neil Houlsby.
\newblock An image is worth 16x16 words: Transformers for image recognition at
  scale.
\newblock In \emph{International Conference on Learning Representations
  (ICLR)}, 2021.

\bibitem[Frankle et~al.(2020)Frankle, Dziugaite, Roy, and
  Carbin]{frankle2020linear}
Jonathan Frankle, Gintare~Karolina Dziugaite, Daniel Roy, and Michael Carbin.
\newblock Linear mode connectivity and the lottery ticket hypothesis.
\newblock In \emph{International Conference on Machine Learning (ICML)}, 2020.

\bibitem[Goyal et~al.(2017)Goyal, Doll{\'a}r, Girshick, Noordhuis, Wesolowski,
  Kyrola, Tulloch, Jia, and He]{goyal2017accurate}
Priya Goyal, Piotr Doll{\'a}r, Ross Girshick, Pieter Noordhuis, Lukasz
  Wesolowski, Aapo Kyrola, Andrew Tulloch, Yangqing Jia, and Kaiming He.
\newblock Accurate, large minibatch sgd: Training imagenet in 1 hour.
\newblock \emph{arXiv preprint arXiv:1706.02677}, 2017.

\bibitem[{Gressmann} et~al.(2020){Gressmann}, {Eaton-Rosen}, and
  {Luschi}]{gressmann2020improving}
Frithjof {Gressmann}, Zach {Eaton-Rosen}, and Carlo {Luschi}.
\newblock Improving neural network training in low dimensional random bases.
\newblock In \emph{Advances in Neural Information Processing Systems
  (NeurIPS)}, 2020.

\bibitem[Gupta et~al.(2019)Gupta, Serrano, and DeCoste]{gupta2019stochastic}
Vipul Gupta, Santiago~Akle Serrano, and Dennis DeCoste.
\newblock Stochastic weight averaging in parallel: Large-batch training that
  generalizes well.
\newblock In \emph{International Conference on Learning Representations
  (ICLR)}, 2019.

\bibitem[Gur-Ari et~al.(2018)Gur-Ari, Roberts, and Dyer]{gur2018gradient}
Guy Gur-Ari, Daniel~A Roberts, and Ethan Dyer.
\newblock Gradient descent happens in a tiny subspace.
\newblock \emph{arXiv preprint arXiv:1812.04754}, 2018.

\bibitem[He et~al.(2016)He, Zhang, Ren, and Sun]{he2016deep}
Kaiming He, Xiangyu Zhang, Shaoqing Ren, and Jian Sun.
\newblock Deep residual learning for image recognition.
\newblock In \emph{Proceedings of the IEEE Conference on Computer Vision and
  Pattern Recognition (CVPR)}, pages 770--778, 2016.

\bibitem[Hu et~al.(2022)Hu, Shen, Wallis, Allen-Zhu, Li, Wang, Wang, and
  Chen]{hu2022lora}
Edward~J Hu, Yelong Shen, Phillip Wallis, Zeyuan Allen-Zhu, Yuanzhi Li, Shean
  Wang, Lu~Wang, and Weizhu Chen.
\newblock Lo{RA}: Low-rank adaptation of large language models.
\newblock In \emph{International Conference on Learning Representations
  (ICLR)}, 2022.

\bibitem[Izmailov et~al.(2018)Izmailov, Podoprikhin, Garipov, Vetrov, and
  Wilson]{izmailov2018averaging}
Pavel Izmailov, Dmitrii Podoprikhin, Timur Garipov, Dmitry Vetrov, and
  Andrew~Gordon Wilson.
\newblock Averaging weights leads to wider optima and better generalization.
\newblock \emph{arXiv preprint arXiv:1803.05407}, 2018.

\bibitem[Kaddour(2022)]{kaddour2022stop}
Jean Kaddour.
\newblock Stop wasting my time! saving days of imagenet and bert training with
  latest weight averaging.
\newblock \emph{arXiv preprint arXiv:2209.14981}, 2022.

\bibitem[{Kingma} and {Ba}(2015)]{kingma2014adam}
Diederik~P. {Kingma} and Jimmy~Lei {Ba}.
\newblock Adam: A method for stochastic optimization.
\newblock In \emph{International Conference on Learning Representations
  (ICLR)}, 2015.

\bibitem[Krizhevsky and Hinton(2009)]{krizhevsky2009learning}
Alex Krizhevsky and Geoffrey Hinton.
\newblock Learning multiple layers of features from tiny images.
\newblock \emph{Technical Report}, 2009.

\bibitem[{Li} et~al.(2018){Li}, {Farkhoor}, {Liu}, and
  {Yosinski}]{li2018measuring}
Chunyuan {Li}, Heerad {Farkhoor}, Rosanne {Liu}, and Jason {Yosinski}.
\newblock Measuring the intrinsic dimension of objective landscapes.
\newblock In \emph{International Conference on Learning Representations
  (ICLR)}, 2018.

\bibitem[Li et~al.(2020)Li, Zhao, Varma, Salpekar, Noordhuis, Li, Paszke,
  Smith, Vaughan, Damania, et~al.]{li2020pytorch}
Shen Li, Yanli Zhao, Rohan Varma, Omkar Salpekar, Pieter Noordhuis, Teng Li,
  Adam Paszke, Jeff Smith, Brian Vaughan, Pritam Damania, et~al.
\newblock Pytorch distributed: Experiences on accelerating data parallel
  training.
\newblock \emph{arXiv preprint arXiv:2006.15704}, 2020.

\bibitem[Li et~al.(2022{\natexlab{a}})Li, Tan, Huang, Tao, Liu, and
  Huang]{li2022low}
Tao Li, Lei Tan, Zhehao Huang, Qinghua Tao, Yipeng Liu, and Xiaolin Huang.
\newblock Low dimensional trajectory hypothesis is true: Dnns can be trained in
  tiny subspaces.
\newblock \emph{IEEE Transactions on Pattern Analysis and Machine Intelligence
  (TPAMI)}, 2022{\natexlab{a}}.

\bibitem[Li et~al.(2022{\natexlab{b}})Li, Wu, Chen, Fang, and
  Huang]{li2022subspace}
Tao Li, Yingwen Wu, Sizhe Chen, Kun Fang, and Xiaolin Huang.
\newblock Subspace adversarial training.
\newblock In \emph{Proceedings of the IEEE Conference on Computer Vision and
  Pattern Recognition (CVPR)}, 2022{\natexlab{b}}.

\bibitem[Li et~al.(2023)Li, Huang, Tao, Wu, and Huang]{li2023trainable}
Tao Li, Zhehao Huang, Qinghua Tao, Yingwen Wu, and Xiaolin Huang.
\newblock Trainable weight averaging: Efficient training by optimizing
  historical solutions.
\newblock In \emph{International Conference on Learning Representations
  (ICLR)}, 2023.

\bibitem[Merity et~al.(2016)Merity, Xiong, Bradbury, and
  Socher]{merity2016pointer}
Stephen Merity, Caiming Xiong, James Bradbury, and Richard Socher.
\newblock Pointer sentinel mixture models.
\newblock \emph{arXiv preprint arXiv:1609.07843}, 2016.

\bibitem[Nesterov(1983)]{nesterov1983method}
Yurii~E Nesterov.
\newblock A method for solving the convex programming problem with convergence
  rate o (1/k\^{} 2).
\newblock In \emph{Dokl. akad. nauk Sssr}, volume 269, pages 543--547, 1983.

\bibitem[Paszke et~al.(2017)Paszke, Gross, Chintala, Chanan, Yang, DeVito, Lin,
  Desmaison, Antiga, and Lerer]{paszke2017automatic}
Adam Paszke, Sam Gross, Soumith Chintala, Gregory Chanan, Edward Yang, Zachary
  DeVito, Zeming Lin, Alban Desmaison, Luca Antiga, and Adam Lerer.
\newblock Automatic differentiation in pytorch.
\newblock 2017.

\bibitem[Polyak and Juditsky(1992)]{polyak1992acceleration}
Boris~T Polyak and Anatoli~B Juditsky.
\newblock Acceleration of stochastic approximation by averaging.
\newblock \emph{SIAM journal on control and optimization}, 30\penalty0
  (4):\penalty0 838--855, 1992.

\bibitem[Rabenseifner(2004)]{rabenseifner2004optimization}
Rolf Rabenseifner.
\newblock Optimization of collective reduction operations.
\newblock In \emph{International Conference on Computational Science}, pages
  1--9. Springer, 2004.

\bibitem[Radford et~al.(2019)Radford, Wu, Child, Luan, Amodei, Sutskever,
  et~al.]{radford2019language}
Alec Radford, Jeffrey Wu, Rewon Child, David Luan, Dario Amodei, Ilya
  Sutskever, et~al.
\newblock Language models are unsupervised multitask learners.
\newblock \emph{OpenAI blog}, 1\penalty0 (8):\penalty0 9, 2019.

\bibitem[Radford et~al.(2021)Radford, Kim, Hallacy, Ramesh, Goh, Agarwal,
  Sastry, Askell, Mishkin, Clark, et~al.]{radford2021learning}
Alec Radford, Jong~Wook Kim, Chris Hallacy, Aditya Ramesh, Gabriel Goh,
  Sandhini Agarwal, Girish Sastry, Amanda Askell, Pamela Mishkin, Jack Clark,
  et~al.
\newblock Learning transferable visual models from natural language
  supervision.
\newblock In \emph{International Conference on Machine Learning (ICML)}, 2021.

\bibitem[Rame et~al.(2022)Rame, Kirchmeyer, Rahier, Rakotomamonjy, Gallinari,
  and Cord]{rame2022diverse}
Alexandre Rame, Matthieu Kirchmeyer, Thibaud Rahier, Alain Rakotomamonjy,
  Patrick Gallinari, and Matthieu Cord.
\newblock Diverse weight averaging for out-of-distribution generalization.
\newblock In \emph{Advances in Neural Information Processing Systems
  (NeurIPS)}, 2022.

\bibitem[Sanyal et~al.(2023)Sanyal, Neerkaje, Kaddour, Kumar, and
  Sanghavi]{sanyal2023early}
Sunny Sanyal, Atula Neerkaje, Jean Kaddour, Abhishek Kumar, and Sujay Sanghavi.
\newblock Early weight averaging meets high learning rates for llm
  pre-training.
\newblock \emph{arXiv preprint arXiv:2306.03241}, 2023.

\bibitem[Shen et~al.(2023)Shen, Sun, Yu, Ding, Tian, and
  Tao]{shen2023efficient}
Li~Shen, Yan Sun, Zhiyuan Yu, Liang Ding, Xinmei Tian, and Dacheng Tao.
\newblock On efficient training of large-scale deep learning models: A
  literature review.
\newblock \emph{arXiv preprint arXiv:2304.03589}, 2023.

\bibitem[Shen et~al.(2020)Shen, Dong, Ye, Ma, Yao, Gholami, Mahoney, and
  Keutzer]{shen2020q}
Sheng Shen, Zhen Dong, Jiayu Ye, Linjian Ma, Zhewei Yao, Amir Gholami,
  Michael~W Mahoney, and Kurt Keutzer.
\newblock Q-bert: Hessian based ultra low precision quantization of bert.
\newblock In \emph{Proceedings of the AAAI Conference on Artificial
  Intelligence (AAAI)}, 2020.

\bibitem[Simonyan and Zisserman(2014)]{simonyan2014very}
Karen Simonyan and Andrew Zisserman.
\newblock Very deep convolutional networks for large-scale image recognition.
\newblock \emph{arXiv preprint arXiv:1409.1556}, 2014.

\bibitem[Tuddenham et~al.(2020)Tuddenham, Pr{\"u}gel-Bennett, and
  Hare]{tuddenham2020quasi}
Mark Tuddenham, Adam Pr{\"u}gel-Bennett, and Jonathan Hare.
\newblock Quasi-newton's method in the class gradient defined high-curvature
  subspace.
\newblock \emph{arXiv preprint arXiv:2012.01938}, 2020.

\bibitem[Vaswani et~al.(2017)Vaswani, Shazeer, Parmar, Uszkoreit, Jones, Gomez,
  Kaiser, and Polosukhin]{vaswani2017attention}
Ashish Vaswani, Noam Shazeer, Niki Parmar, Jakob Uszkoreit, Llion Jones,
  Aidan~N Gomez, {\L}ukasz Kaiser, and Illia Polosukhin.
\newblock Attention is all you need.
\newblock In \emph{Advances in Neural Information Processing Systems
  (NeurIPS)}, 2017.

\bibitem[Vinyals and Povey(2012)]{vinyals2012krylov}
Oriol Vinyals and Daniel Povey.
\newblock Krylov subspace descent for deep learning.
\newblock In \emph{Artificial Intelligence and Statistics (AISTATS)}, pages
  1261--1268. PMLR, 2012.

\bibitem[Wortsman et~al.(2022)Wortsman, Ilharco, Gadre, Roelofs, Gontijo-Lopes,
  Morcos, Namkoong, Farhadi, Carmon, Kornblith, et~al.]{wortsman2022model}
Mitchell Wortsman, Gabriel Ilharco, Samir~Ya Gadre, Rebecca Roelofs, Raphael
  Gontijo-Lopes, Ari~S Morcos, Hongseok Namkoong, Ali Farhadi, Yair Carmon,
  Simon Kornblith, et~al.
\newblock Model soups: averaging weights of multiple fine-tuned models improves
  accuracy without increasing inference time.
\newblock In \emph{International Conference on Machine Learning (ICML)}, 2022.

\bibitem[Yang et~al.(2019)Yang, Zhang, Kirichenko, Bai, Wilson, and
  De~Sa]{yang2019swalp}
Guandao Yang, Tianyi Zhang, Polina Kirichenko, Junwen Bai, Andrew~Gordon
  Wilson, and Chris De~Sa.
\newblock Swalp: Stochastic weight averaging in low precision training.
\newblock In \emph{International Conference on Machine Learning (ICML)}. PMLR,
  2019.

\bibitem[You et~al.(2020{\natexlab{a}})You, Li, Xu, Fu, Wang, Chen, Baraniuk,
  Wang, and Lin]{you2019drawing}
Haoran You, Chaojian Li, Pengfei Xu, Yonggan Fu, Yue Wang, Xiaohan Chen,
  Richard~G Baraniuk, Zhangyang Wang, and Yingyan Lin.
\newblock Drawing early-bird tickets: Towards more efficient training of deep
  networks.
\newblock In \emph{International Conference on Learning Representations
  (ICLR)}, 2020{\natexlab{a}}.

\bibitem[You et~al.(2017)You, Gitman, and Ginsburg]{you2017large}
Yang You, Igor Gitman, and Boris Ginsburg.
\newblock Large batch training of convolutional networks.
\newblock \emph{arXiv preprint arXiv:1708.03888}, 2017.

\bibitem[You et~al.(2020{\natexlab{b}})You, Li, Reddi, Hseu, Kumar,
  Bhojanapalli, Song, Demmel, Keutzer, and Hsieh]{You2020Large}
Yang You, Jing Li, Sashank Reddi, Jonathan Hseu, Sanjiv Kumar, Srinadh
  Bhojanapalli, Xiaodan Song, James Demmel, Kurt Keutzer, and Cho-Jui Hsieh.
\newblock Large batch optimization for deep learning: Training bert in 76
  minutes.
\newblock In \emph{International Conference on Learning Representations
  (ICLR)}, 2020{\natexlab{b}}.

\bibitem[Zhang et~al.(2019)Zhang, Lucas, Ba, and Hinton]{zhang2019lookahead}
Michael Zhang, James Lucas, Jimmy Ba, and Geoffrey~E Hinton.
\newblock Lookahead optimizer: k steps forward, 1 step back.
\newblock In \emph{Advances in Neural Information Processing Systems
  (NeurIPS)}, 2019.

\bibitem[Zhang et~al.(2024)Zhang, Jaiswal, Yin, Liu, Zhao, Tian, and
  Wang]{zhang2024q}
Zhenyu Zhang, Ajay Jaiswal, Lu~Yin, Shiwei Liu, Jiawei Zhao, Yuandong Tian, and
  Zhangyang Wang.
\newblock Q-galore: Quantized galore with int4 projection and layer-adaptive
  low-rank gradients.
\newblock \emph{arXiv preprint arXiv:2407.08296}, 2024.

\bibitem[Zhou et~al.(2017)Zhou, Yao, Guo, Xu, and Chen]{zhou2017incremental}
Aojun Zhou, Anbang Yao, Yiwen Guo, Lin Xu, and Yurong Chen.
\newblock Incremental network quantization: Towards lossless cnns with
  low-precision weights.
\newblock \emph{arXiv preprint arXiv:1702.03044}, 2017.

\end{thebibliography}

\appendix
\section{Training Details}
\label{sec:training_details}
In the following, we list the exact training hyper-parameters used for our experiments.
\begin{table}[htbp]
    \centering
    \caption{Training hyper-parameters on CIFAR-100.}
    \vspace{2mm}
    \label{tab:hp-cifar-100}
    \begin{tabular}{lcc}
    \toprule
    Models &VGG / ResNet &ViTs \\
    \midrule
         Base Optimizer &SGD &AdamW  \\
        Epochs &\multicolumn{2}{c}{200}  \\
        Warm Up Epochs &\multicolumn{2}{c}{8} \\
        Data Augmentation &\multicolumn{2}{c}{Inception style} \\
        Peak Learning Rate &0.1 &1e-3 \\
        LR-Scheduler &Step &Cosine \\
        Batch Size &\multicolumn{2}{c}{128} \\
        Weight Decay &0.0001 &0.1 \\
        \midrule
        \multicolumn{3}{c}{TWA Hyper-parameters} \\
        Optimizer &SGD &AdamW \\ 
        Peak Learning Rate  &\multicolumn{2}{c}{0.01} \\
        LR-Scheduler &\multicolumn{2}{c}{Cosine} \\
        Weight Decay &\multicolumn{2}{c}{0} \\
    \bottomrule
    \end{tabular}
\end{table}

\begin{table}[htbp]
    \centering
    \caption{Training hyper-parameters on ImageNet.}
    \vspace{2mm}
    \label{tab:hp-imagenet}
    \begin{tabular}{lcc}
    \toprule
    Models &ResNets &ViTs \\
    \midrule
         Base Optimizer &SGD &AdamW  \\
        Epochs &\multicolumn{2}{c}{90}  \\
        Warm Up Epochs &\multicolumn{2}{c}{8} \\
        Data Augmentation &\multicolumn{2}{c}{Inception style} \\
        Peak Learning Rate &0.4 &1e-3 \\
        LR-Scheduler &Step &Cosine \\
        Batch Size &\multicolumn{2}{c}{1024} \\
        Weight Decay &0.0001 &0.1 \\  
        \midrule
        \multicolumn{3}{c}{TWA Hyper-parameters} \\
        Optimizer &SGD &AdamW \\ 
        Peak Learning Rate  &\multicolumn{2}{c}{0.01} \\
        LR-Scheduler &\multicolumn{2}{c}{Cosine} \\
        Weight Decay &\multicolumn{2}{c}{0} \\
    \bottomrule
    \end{tabular}
\end{table}

\begin{table}[htbp]
    \centering
    \caption{Training hyper-parameters on  WMT2014.}
    \vspace{2mm}
    \label{tab:hp-imagenet}
    \begin{tabular}{lc}
    \toprule
    Models &Transformer \\
    \midrule
         Base Optimizer &Adam  \\
        Steps &\multicolumn{1}{c}{100k}  \\
        Peak Learning Rate &0.0001 \\
        LR-Scheduler &ReduceLROnPlateau \\
        Batch Size &\multicolumn{1}{c}{256} \\
        Dropout &0.1 \\
        Weight Decay &0.0001 \\  
        \midrule
        \multicolumn{2}{c}{TWA Hyper-parameters} \\
        Optimizer &Adam \\ 
        Peak Learning Rate  &\multicolumn{1}{c}{0.01} \\
        LR-Scheduler &\multicolumn{1}{c}{Constant} \\
        Weight Decay &\multicolumn{1}{c}{0} \\
    \bottomrule
    \end{tabular}
\end{table}

\begin{table}[htbp]
    \centering
    \caption{Training hyper-parameters on  CIFAR-10 and ImageNet with CLIP ViT-B/32.}
    \vspace{2mm}
    \label{tab:hp-imagenet}
    \begin{tabular}{lc}
    \toprule
    Models &CLIP ViT-B/32 \\
    \midrule
    
        \multicolumn{2}{c}{TWA Hyper-parameters} \\
        Batch Size &128 \\
        Optimizer &AdamW \\ 
        Peak Learning Rate  &\multicolumn{1}{c}{0.01} \\
        Weight Decay &0 \\
        LR-Scheduler &\multicolumn{1}{c}{Cosine} \\
    \bottomrule
    \end{tabular}
\end{table}

\begin{table}[htbp]
    \centering
    \caption{Training hyper-parameters on   WikiText-2  with GPT-2.}
    \vspace{2mm}
    \label{tab:hp-wikitext}
    \begin{tabular}{lc}
    \toprule
    Models &GPT-2 \\
    \midrule
    Base Optimizer &AdamW  \\
        Steps &\multicolumn{1}{c}{1,000}  \\
        Peak Learning Rate &0.00005 \\
        LR-Scheduler &Linear \\
        Batch Size &\multicolumn{1}{c}{8} \\
        Weight Decay &0 \\  
        \midrule
        \multicolumn{2}{c}{TWA Hyper-parameters} \\
        Batch Size &8 \\
        Optimizer &AdamW \\ 
        Peak Learning Rate  &\multicolumn{1}{c}{0.01} \\
        Weight Decay &0 \\
        LR-Scheduler &\multicolumn{1}{c}{Cosine} \\
    \bottomrule
    \end{tabular}
\end{table}

\newpage
\section{Dataset Details}
We provide a detailed overview of the dataset splits used in this paper in Table~\ref{tab:datasets}.
\begin{table}[htbp]
    \centering
    \caption{Datasets splitting information.}
    \vspace{2mm}
    \label{tab:datasets}
    \begin{tabular}{c|ccc}
         Datasets & Training &Validation &Test \\
         \hline
         CIFAR-10/100 &45,000 &5,000 &10,000  \\
         ImageNet &1,255,167 &26,000 &50,000 \\
         WMT2014 &954,000 &3,000 &3,000 \\
         Wikitext-2 &36718 &3760 &4358 
    \end{tabular}
\end{table}


\end{document}